\documentclass[conference]{IEEEtran}
\usepackage{times}
\let\labelindent\relax

\usepackage[numbers]{natbib}
\usepackage{multicol}
\usepackage[bookmarks=true]{hyperref}
\usepackage{amssymb}

\def\CircleArrowright{\ensuremath{%
  \rotatebox[origin=c]{310}{$\circlearrowright$}}}
\newcommand{\vlnbert}{VLN$\protect\CircleArrowright$BERT}

\usepackage{times}

\usepackage[numbers]{natbib}
\usepackage{multicol}
\usepackage[bookmarks=true]{hyperref}
\usepackage{graphicx}
\usepackage{url}            
\usepackage{booktabs}       
\usepackage{amsfonts}       
\usepackage{nicefrac}       
\usepackage{microtype}      
\usepackage{xcolor}         
\usepackage{graphicx}
\usepackage{makecell}
\usepackage{multirow}
\usepackage{floatrow}
\usepackage{wrapfig}
\usepackage{soul}
\usepackage{amsmath}
\usepackage{subcaption}
\usepackage{wrapfig,lipsum,booktabs}
\usepackage{xcolor,colortbl}
\usepackage{svg}
\usepackage{bbding}
\usepackage{enumitem}

\usepackage{booktabs} 
\usepackage{adjustbox} 
\usepackage{tabularx} 

\usepackage{acronym}

\usepackage{tcolorbox}
\usepackage{algorithm}
\usepackage{algpseudocode}

\definecolor{mygreen}{rgb}{0.643, 0.784, 0.533}
\definecolor{myblue}{rgb}{0.521,0.631,0.831}

\newcommand{\name}{Uni-NaVid}

\newcommand{\egno}{\textit{e}.\textit{g}.} 

\newcommand{\cI}{\mathcal{I}}
\newcommand{\cO}{\mathcal{O}}

\newcommand{\ccomment}[1]{\textcolor[rgb]{0.180, 0.372, 0.498}{#1}}

\makeatletter
\def\blfootnote{\xdef\@thefnmark{}\@footnotetext}
\makeatother

\begin{document}

\title{\name: A Video-based Vision-Language-Action Model for Unifying Embodied Navigation Tasks}


\author{Jiazhao Zhang$^{1,2}$ \quad\quad Kunyu Wang$^{3}$ \quad\quad Shaoan Wang$^{1,2}$ \quad\quad Minghan Li$^{2}$ \quad\quad  Haoran Liu$^{1,2}$ \quad\quad \\ Songlin Wei$^{1,2}$  \quad\quad  Zhongyuan Wang$^{3}$ \quad\quad Zhizheng Zhang$^{2,3,\dagger}$ \quad\quad He Wang$^{1,2,3,\dagger}$ \\ \normalsize{ $^1$CFCS, School of Computer Science, Peking University  \quad $^2$Galbot \quad $^3$Beijing Academy of Artificial Intelligence} \\ \href{https://pku-epic.github.io/Uni-NaVid}{\texttt{https://pku-epic.github.io/Uni-NaVid} }}



%


\twocolumn[{%
\renewcommand\twocolumn[1][]{#1}%
\maketitle
\vspace{-0.5cm}
\begin{center}
    \centering
    \captionsetup{type=figure}
    \includegraphics[width=1\linewidth]{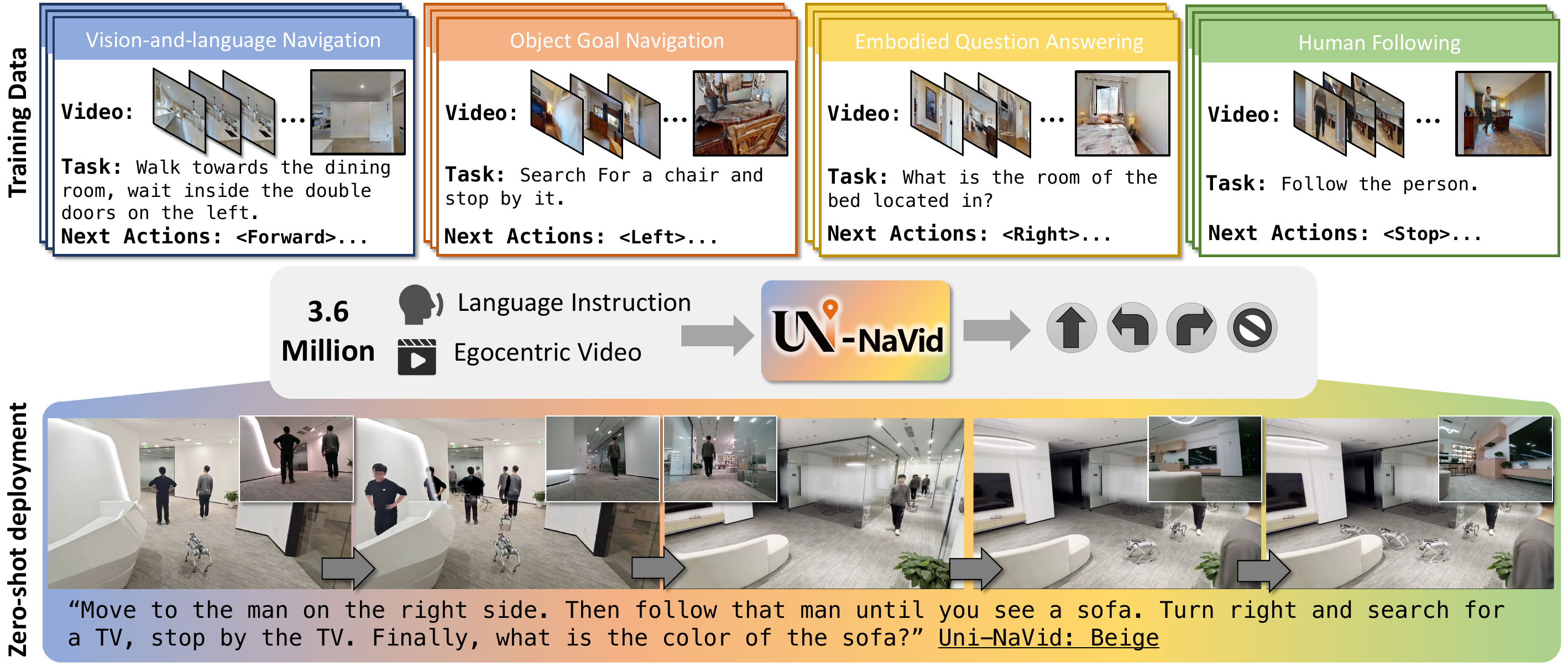}
    \captionof{figure}{\name~learns general navigation skills across four embodied navigation tasks using 3.6 million navigation samples. \name~only takes online RGB video frames and language instructions as input and output actions, achieving general navigation ability in a real-world deployment.   }
    \label{fig:teaser}
\end{center}
}]

\blfootnote{ Author (e-mail: zhngjizh@gmail.com). $\dagger$ Corresponding authors (e-mail: zhangzz@galbot.com, hewang@pku.edu.cn).}


\begin{abstract}

Embodied Navigation is a fundamental capability for intelligent robots, requiring robots to follow human commands and move autonomously within physical environments. Despite significant advancements, most existing navigation approaches are tailored to specific navigation tasks, such as instruction following, searching objects, answering questions, tracking people, and more. However, the increasing demands on advanced embodied navigation pose the challenge of designing a practical navigation agent that can incorporate multiple navigation tasks naturally and benefits from the synergy between these tasks. 
To this end, we present \name, a video-based vision-language-action (VLA) model to unify different paradigms of navigation tasks and improve navigation performance by encouraging the synergy among different navigation sub-tasks. This VLA model can directly take natural language instructions and RGB video streams as inputs and output low-level robotic actions in an end-to-end manner.  To efficiently process extensive RGB video streams, we propose an online token merge strategy that spatially and temporally consolidates similar visual information which improves the inference speed to 5 Hz.
For training \name, we collect 3.6~million navigation data samples across different navigation tasks. Extensive experiments on diverse navigation benchmarks demonstrate that \name~achieves state-of-the-art performance within a unified framework by using only ego-centric RGB video as inputs. Additionally, real-world experiments confirm the model’s effectiveness and efficiency, shedding light on its strong generalizability. 

\end{abstract}

\IEEEpeerreviewmaketitle

\section{Introduction}
\label{sec:intro}



Embodied navigation~\cite{Zhang2024VisionandLanguageNT,wang2022towards} is a critical capability for intelligent robots and has drawn significant attention in the robotics community. For successful embodied navigation, robots must be able to move autonomously within physical environments based on human instructions. However, navigation tasks vary significantly, and most existing studies are designed for specific tasks, e.g., vision-and-language navigation~\cite{Krantz2020BeyondTN, anderson2020rxr}, object goal navigation~\cite{chaplot2020object}, embodied question answering~\cite{das2018embodied, wijmans2019embodied}, and following~\cite{zhang2021efficient, huang2017robust, puig2023habitat}. Consequently, most current approaches are developed to address only one type of navigation task, often relying on specialized modules and task-specific datasets. This narrow scope limits their applicability to multi-purpose navigation applications and prevents these methods from leveraging potential synergies across diverse navigation tasks.

Developing a versatile navigation model presents significant challenges, as it requires the unification of navigation task modeling and the integration of heterogeneous data for joint use. Initial efforts adopt imitation learning (IL)~\cite{wang2022towards,wu2020towards, nguyen2019vision} or reinforcement learning (RL)~\cite{zeng2024poliformer, xu2023benchmarking} to learn general navigation skills in simulation environments or limited diverse real-world environments. 
However, due to the limited rendering quality and diversity of simulators, these approaches often encounter the ``sim-to-real'' gap and suffer from poor generalization across diverse navigation tasks~\cite{gervet2023navigating, anderson2021sim, kadian2020sim2real}.
Recent studies~\cite{zhou2023navgpt, zheng2023towards, long2024instructnav, long2023discuss, shah2023lm} have attempted to achieve a higher degree of unification using pre-trained large language models (LLMs). However, due to the low frequency of LLM inference, they simplify the problem to some extent by adopting discretized modeling approaches. They rely on pre-defined graphs for decision-making learning, which sacrifices output flexibility and introduces additional challenges for real-world deployment.

In this work, we propose \name, a video-based Vision-Language-Action (VLA) model for unifying diverse commonly demanded navigation tasks (Tab.~\ref{tab:comp-task}). \name~ takes egocentric RGB video streams and natural language instructions as inputs, and directly generates low-level actions for navigation in continuous environments. To achieve multi-task navigation while supporting efficient navigation, \name{} extend video-based VLM~\cite{li2023llama} by incoprating two key components: (1) an efficient VLA architecture based on an online token merge mechanism, which enables efficient processing of online-captured video streams for LLM inference; and (2) an extensive collection of 3.6M samples across four widely studied navigation tasks. We provide a detailed elaboration below:


During navigation, the agent is required to process a substantial volume of online captured frames, which results in memory overload and computational latency, particularly in LLM-based approaches~\cite{zhang2024navid, long2024instructnav}. To this end, we propose an online token merging mechanism to compress near historical frames with a relatively low ratio while compressing far historical frames with a relatively high ratio. This merging mechanism operates in an on-the-fly manner, maximizing the reuse of previous navigation history. In this way, \name{} learn compact representations that maintain not only fine-grained spatial information but also structured temporal information, thus speeding up the model inference by reducing the token number. Besides, \name~adopts a foresight prediction to generate actions for a future horizon at once instead of step-by-step. This enables \name{} to achieve 5Hz inference, facilitating the deployment of a non-blocking navigation robot powered by a VLA model in real-world environments (Please refer to the supplementary video).






We aim to build ~\name{} as a versatile multi-task navigation agent, incorporating four widely demanded navigation tasks: vision-and-language navigation, object-goal navigation, embodied question answering, and human following. These tasks are distinct from each other, with varying task settings and objectives. Specifically, for the human-following task, we construct a new language-guided human-following benchmark for data collection and evaluation. Finally, we collect 3.6M navigation samples based on diverse navigation tasks with different simulation environments. Additionally, inspired by the success of manipulation VLAs~\cite{brohan2023rt}, we further integrate 2.3M real-world internet data samples for Video Question Answering (VQA)~\cite{azuma2022scanqa,li2023llama} and video captioning~\cite{chen2024panda} as auxiliary tasks. This integration aims to enhance scene understanding and promote sim-to-real generalization.



We conduct extensive experiments on benchmarks across the aforementioned four navigation tasks and compared our method with strong baselines specifically designed for each task. Utilizing only RGB video streams and instructions as inputs, our method demonstrates the superiority of a single VLA model across diverse benchmarks, achieving SOTA or SOTA-comparable performance. Furthermore, comprehensive ablation studies validate the synergistic benefits of learning multiple navigation tasks jointly. Finally, real-world experiments demonstrate that \name{} achieves non-blocking navigation exhibiting impressive robustness in handling diverse instructions and environments. We believe our work serves merely as a starting point for general-purpose navigation, and \textit{we will release the full source code to benefit the community.}





\section{Related Works}
\label{sec:formatting}

\begin{figure*}[ht]
\setlength{\abovecaptionskip}{1pt}
\begin{center}
  \includegraphics[width=1 \linewidth]{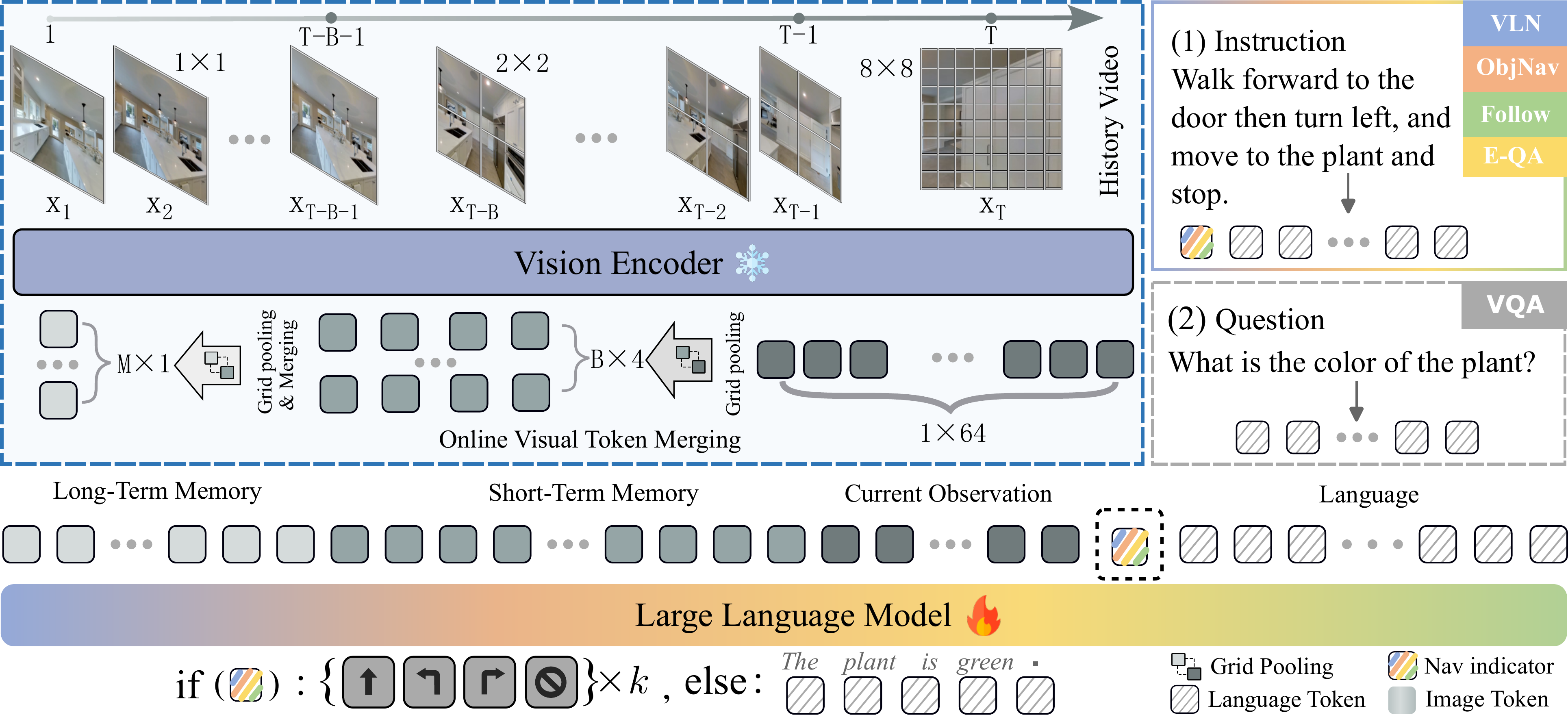}
\end{center}
   \caption{\textbf{Pipeline of \name.} Our method takes only single-view RGB frames $\{\mathbf{x}_1, \cdots, \mathbf{x}_T\}$ and a natural language instruction $\cI$ as input. For each frame, we extract 64 visual tokens using the vision encoder and then use online token merging to accelerate the model while retaining compact visual information. The merged tokens and instruction tokens are sent to the large language model to obtain actions for navigation or answers for embodied question-answering. }
\label{fig:pipeline}
\end{figure*}

\begin{table}[!t]
\centering

\scalebox{0.7}{
\setlength{\tabcolsep}{1mm}{
\begin{tabular}{l|cc|cccc}
\hline
\multirow{2}{*}{Methods} & \multicolumn{2}{c|}{Action} & \multicolumn{4}{c}{Embodied Navigation Tasks} \\
                                       & D.E.         & C.E.         & VLN~\cite{Krantz2020BeyondTN}        & ObjNav~\cite{savva2019habitat}  & EQA~\cite{wijmans2019embodied}        & Follow~\cite{puig2023habitat}  \\ \hline
VLMaps~\cite{huang2022visual}          & \checkmark &            & \checkmark & \checkmark &            &            \\
NaviLLM~\cite{zheng2023towards}        & \checkmark &            & \checkmark & \checkmark & \checkmark &            \\
InstructNav~\cite{long2024instructnav} & \checkmark  &            & \checkmark & \checkmark &            &            \\
Poliformer~\cite{zeng2024poliformer}   &            & \checkmark &            & \checkmark &            & \checkmark \\
\name                             &            & \checkmark & \checkmark & \checkmark & \checkmark & \checkmark \\ \hline
\end{tabular}
}
}
\caption{\textbf{Task and setting comparison. } \name~is developed to address four embodied navigation tasks, generating action outputs in continuous environments. C.E.: Continuous Environment; D.E.: Discrete Environment.}
\label{tab:comp-task}
\end{table} 



\textbf{Multi-Task Embodied Navigation.}
Embodied navigation~\cite{anderson2018evaluation, wu2024embodied, Zhang2024VisionandLanguageNT} requires agents to navigate in unseen environments based on human instructions. There is extensive literature on embodied navigation; here, we focus on four mainstream tasks that involve both visual information and language instructions: Vision-and-Language Navigation~\cite{anderson2018vision, Krantz2020BeyondTN,anderson2020rxr}, Object Goal Navigation~\cite{chaplot2020object}, Embodied Question Answering~\cite{das2018embodied}, and Human Following~\cite{islam2019person, puig2023habitat,zhong2023rspt,zhong2024empowering}. Early efforts~\cite{wang2022towards,wu2020towards,nguyen2019vision,xu2023benchmarking} towards a generalist-embodied navigation model involved multi-task navigation datasets and directly learning navigation skills, showing initial success in multi-task performance. However, these methods experienced performance drops when deployed in novel environments, especially in real-world settings. In recent years, advanced approaches~\cite{zheng2023towards, long2024instructnav, huang2023visual, long2023discuss, zhou2025navgpt,shah2023lm} have leveraged the generalization capabilities of large language models to improve multi-task navigation. These models show promising generalizability across navigation tasks but rely on extensive prompting, which impacts time efficiency. In contrast, our video-based large language model is trained end-to-end for multi-task navigation, offering robust generalization and computational efficiency for tasks like human following.

\noindent\textbf{Embodied Navigation Datasets.} To train and evaluate the performance of a policy for embodied navigation tasks, a large body of datasets and corresponding benchmarks have been proposed \cite{duan2022survey,zhu2021deep,liu2024aligning,mavrogiannis2023core}. These datasets play a crucial role in the embodied navigation community. Here, we review the datasets most relevant to our methods. For vision-and-language navigation, the most widely used datasets are Room-2-Room (R2R) \cite{anderson2018vision} and Room-cross-Room (RxR) \cite{ku2020room}, which provide navigation instructions and ground truth trajectories of landmarks. We focus on a variant of R2R and RxR in continuous environments, called VLN-CE \cite{Krantz2020BeyondTN}, which is more practical for real-world applications. For object goal navigation, there are several famous benchmarks such as HM3D \cite{ramakrishnan2021habitat}, MP3D \cite{chang2017matterport3d}, and Aithor \cite{zhu2017target}, which are built on various scene environments and simulators. Here, we leverages the HM3D dataset on Habitat \cite{savva2019habitat}, which shares the same action settings as VLN-CE. For embodied question answering (EQA), there are diverse datasets focusing on different attributes of EQA, such as MP3D-EQA \cite{wijmans2019embodied}, MT-EQA \cite{yu2019multi}, Graph-EQA \cite{tan2023knowledge}, and MX-EQA \cite{islam2023eqa}. We select MP3D-EQA, which is well-maintained with the latest baselines. For human-following \cite{zhong2019ad,zhong2021towards} benchmarks, there is currently no benchmark that provides textual descriptions of humans. Therefore, we have self-built a textual description-based human-following benchmark using Habitat 3.0 \cite{puig2023habitat}. Note that new benchmarks are consistently being proposed, covering a diverse range of navigation attributes. However, our goal is to train and evaluate our method on mainstream datasets to clearly justify the performance of our approach.

\textbf{Large Language Models for Navigation.}
Large Language Models (LLMs)\cite{chiang2023vicuna, liu2023llava, zhu2023chatgpt} have been introduced into robotic navigation due to their generalization capabilities in understanding and planning. One straightforward approach\cite{zhou2023navgpt, long2024instructnav, long2023discuss,shah2023lm} is to use off-the-shelf large language models in a zero-shot manner. These methods employ visual foundation models~\cite{dorbala2022clip, liu2023llava} to describe surrounding environments in text format, prompting the language model to select landmarks that guide the agent. However, abstracting dense visual information into text and relying on discrete landmarks results in sparse environmental observations and is limited to static environments.
Another approach~\cite{zhang2024navid, zeng2024poliformer} trains a video-based large language model end-to-end with low-level actions to enable continuous movement. However, it faces efficiency challenges in long-horizon tasks. In contrast, \name~implements an online visual token merging strategy, optimizing training efficiency for long-horizon tasks and supporting non-blocking execution in real-world environments.

\section{Problem Formulation}

\textbf{Navigation task definition.} We define the general-purpose navigation of \name~ as follows: At the time $T$, given a natural language instruction $\cI$ consisting of $l$ words and an ego-centric RGB video $\cO_T$ comprising a sequence of frames $\{\mathbf{x}_1, \cdots, \mathbf{x}_T\}$, the agent is required to plan the next $k$ actions $\{\mathcal{A}_{T}, \cdots, \mathcal{A}_{T+k-1}\}$ to executed for complete the instruction within novel environments ($k=4$ in our experiments). Here, we adopt a widely used action setting~\cite{savva2019habitat, chaplot2020object, Krantz2020BeyondTN, das2018embodied}, which require the agent to take low-level actions $\mathbf{a} \in \mathcal{A} $, including $\{\texttt{FORWARD}, \texttt{TURN-LEFT}, \texttt{TURN-RIGHT}, \texttt{STOP}\}$. 
Note that, our task formulation is compatible with existing embodied navigation tasks~\cite{savva2019habitat, chaplot2020object, Krantz2020BeyondTN, das2018embodied}, where the discrete low-level actions~\cite{savva2019habitat, chaplot2020object, Krantz2020BeyondTN, das2018embodied} represent a small rotation (30 degrees) or a forward movement (25 cm), making them flexible to be used in continuous environments such obstacle avoidance. We provide a detailed explanation of how these actions are applied in both synthetic and real-world environments in Sec.~\ref{sec:imp_details}

\textbf{Overview.} As illustrated in Figure~\ref{fig:pipeline}, \name~ is composed of three main components: a vision encoder, an online token merge mechanism and a large language model (LLM). First, the online captured video stream is encoded by the vision encoder (EVA-CLIP~\cite{sun2023eva} in implementation) to extract frame-wise visual features in the form of tokens, which we denote them as visual tokens. The visual tokens are then spatially and temporally merged by leveraging an online token merge mechanism. Next, the merged visual tokens are projected with an MLP projector into a feature space aligned with language tokens, which are referred to as visual observation tokens. As common, the instructions are also tokenized as a set of tokens, known as language observation tokens. Both the visual observation tokens and language observation tokens are concatenated and passed to the Large Language Model (LLM), which infers four action tokens that represent the next four actions.

\section{Model of \name}
\label{sec:model}


\subsection{Observation Encoding.} 

Given the ego-centric video up to time $T$, denoted by $\cO_T=\{\mathbf{x}_1. \cdots, \mathbf{x}_T\}$, we encode the video to a sequence of visual features in the form of tokens. For each frame $\mathbf{x}_t$, we first get its visual feature tokens $\mathbf{X}_t \in \mathbb{R}^{N_x \times C}$ with a vision encoder (EVA-CLIP~\cite{sun2023eva} in implementation), where $N_x$ is the patch number ($N_x$ is set to 256) and $C$ is the embedding dimension. 
\begin{equation}
    \label{equ:project}
    \mathbf{X}_{1:T} = Encoder(\mathbf{x}_{1:T})
\end{equation}

The visual features provide rich information that enables the agent to understand its navigation history and plan subsequent actions. However, during navigation, the progressively increasing number of visual tokens~$(T \times N_x)$ results in progressively longer inference times for the LLM (typically 1–2 seconds per inference)~\cite{zhang2024navid}. This increased latency renders LLM-based navigation impractical for deployment in real-world environments.

\subsection{Online Visual Token Merging}

\label{sec:token_merge}

To reduce the number of visual tokens while preserving sufficient navigation visual information, we design an token merging mechanism. This strategy is based on the key insight that recent observations are more critical for navigation, and that visual information between consecutive frames (temporally) and within neighboring pixels (spatially) may be redundant.

\textbf{Visual token grouping.} Drawing inspiration from the Atkinson-Shiffrin memory model~\cite{atkinson1968human, song2024moviechat}, we categorize visual tokens into current visual tokens $\mathbf{X}_{\text{curr}}$, short-term visual tokens $\mathbf{X}_{\text{short}}$, and long-term visual tokens $\mathbf{X}_{\text{long}}$. 
These visual tokens are grouped based on their timestamps relative to the current frame $T$ and for each group of visual tokens, we apply a grid pooling operation at different pooling resolutions:

\vspace{-10pt}
\begin{equation}
\label{equ:visual_token}
    \mathbf{X}_{1:T} = 
\begin{cases}
    \mathbf{X}_{\text{curr}} = GridPool(\mathbf{X}_t, \alpha_\text{curr}), & \text{if t = T} \\
    \mathbf{X}_{\text{short}}= GridPool(\mathbf{X}_t,  \alpha_\text{short}), & \text{if t $\in$ [T-B, T)} \\
    \mathbf{X}_{\text{long}} = GridPool(\mathbf{X}_t, \alpha_\text{long}), & \text{if t $\in$ [1, T-B)}
\end{cases}
\end{equation}

where GridPool(·) is a grid pooling operation ~\cite{li2023llama, zhang2024navid}, \textit{spatially} squeezing the tokens from $N_x$ to $\frac{N_x}{\alpha^2}$, and $B$ (set to $64$) is the length of the buffer of shorter memory. Here, we adopt the $\alpha_\text{curr}=2$,  $\alpha_\text{short}=8$,  $\alpha_\text{long}=16$, leads to visual tokens as  $\mathbf{X}_\text{curr} \in \mathbb{R}^{64\times C}$,  $\mathbf{X}_\text{short} \in \mathbb{R}^{4\times C}$,  $\mathbf{X}_\text{long} \in \mathbb{R}^{1\times C}$, respectively. Here, current visual tokens $\mathbf{X}_\text{curr}$ encapsulate comprehensive visual information, enabling the agent to perceive its immediate environment and plan subsequent trajectories. Meanwhile, $\mathbf{X}_\text{short}$ and $\mathbf{X}_\text{long}$ capture temporally rich information from the captured video stream, facilitating the agent's comprehension of its navigation history. 

It should be noted that these hyperparameters are obtained through empirical experimentation to achieve an optimal balance between manageable token numbers and adequate visual information representation. These hyperparameters can be further adjusted when memory capacity and computational resources are not limiting factors. We provide a detailed explanation and ablation study of $\alpha$ in the supplemental material.

\textbf{Online visual token process.} During the navigation process, the agent consistently observes new frames. However, performing encoding and grouping (Eq.~\ref{equ:visual_token}) for all frames at each step would be computationally intensive. To address this, we implement an online visual token processing mechanism that maximizes the reuse of previously generated visual tokens. Specifically, when a new frame at time $T+1$ is received, we apply grid pooling exclusively to the most recent visual tokens at time $T$ and the oldest short-term visual tokens at time $T-B$. These processed tokens are then integrated into the short-term and long-term visual tokens, respectively:

\begin{align}
    \mathbf{X}_{\text{curr}\rightarrow\text{short}} &= GridPool(\mathbf{X}_\text{curr},\frac{\alpha_\text{short}}{\alpha_\text{curr}}),\\
    \mathbf{X}_{\text{short}\rightarrow\text{long}} &= GridPool(\mathbf{X}_\text{short},\frac{\alpha_\text{long}}{\alpha_\text{short}}).
\end{align}

To prevent the linear growth of long-term visual tokens $\mathbf{X}_\text{Long}$, we further perform token merging on the long-term visual tokens by combining adjacent tokens that exhibit high similarity, following the approach of VLM-based methods~\cite{Bolya2022TokenMY, song2024moviechat}. Specifically, we merge the long-term visual tokens based on the cosine similarity between $ \mathbf{X}_{\text{short}\rightarrow\text{long}}$  and the most recent long-term visual tokens $\mathbf{X}_{\text{long}}$ at time $T-B-1$. If the similarity exceeds a predefined threshold $\tau$, we merge them according to the number of frames previously merged (denoted as $K$) in the latest long-term visual tokens:



\begin{align}
\mathbf{X}_{\text{long}} &= \frac{1}{K+1} \left( K \mathbf{X}_{\text{long}} + \mathbf{X}_{\text{short} \to \text{long}} \right),  \\
\textit{subject to} \quad & \cos \left( \mathbf{X}_{\text{long}}, \mathbf{X}_{\text{short} \to \text{long}} \right)  > \tau.
\end{align}

We insert new long-term visual tokens $\mathbf{X}_{\text{short} \to \text{long}} $ when their similarity falls below a threshold  $\tau$ (empirically set to $\tau = 0.95$~\cite{song2024moviechat}),  indicating that they contain relatively distinct visual information. This online visual token processing preserves the navigation visual history in a highly compact form (with a length of $M \ll T-B-1$). Notably, only visual tokens at the boundaries of groups require parallelizable grid pooling, making the process computationally efficient and naturally suited for online deployment in real-world navigation tasks. We give a description of our token merging technique at Algorithmn~\ref{algo:token-merging}.

\begin{algorithm}[t]
\caption{Online Visual Token Merging}
\label{algo:token-merging}
\begin{algorithmic}[1]
\Require 
\Statex
\begin{itemize}
    \item Total number of frames $T$
    \item Short memory buffer length $B$
    \item Grid pooling scales: $\alpha_{\text{curr}}$, $\alpha_{\text{short}}$, $\alpha_{\text{long}}$
    \item Current visual tokens: $\mathbf{X}_T \in \mathbb{R}^{N_x \times C}$
    \item Previously merged tokens: $\mathbf{X}_{\text{curr}}$, $\mathbf{X}_{\text{short}}$, $\mathbf{X}_{\text{long}}$
    \item Number of frames merged in the last tokens of long memory: $K$
\end{itemize}
\Ensure 
\Statex
\begin{itemize}
    \item Updated merged tokens: $\mathbf{X}'_{\text{curr}}$, $\mathbf{X}'_{\text{short}}$, $\mathbf{X}'_{\text{long}}$
    \item Updated number of frames merged in the last tokens of long memory: $K'$
\end{itemize}
\Statex 

\If{$T == 1$} \Comment{\ccomment{First frame, empty history tokens}}
    \State $\mathbf{X}'_{\text{short}},\mathbf{X}'_{\text{long}}  \gets []$
\Else \Comment{\ccomment{Update short-term visual tokens}}
    \State $\mathbf{X}_{\text{curr}\rightarrow\text{short}} \gets GridPool(\mathbf{X}_\text{curr}, \frac{\alpha_{\text{short}}}{\alpha_{\text{curr}}})$
    \State $\mathbf{X}'_{\text{short}} \gets \mathbf{X}_{\text{short}} + [\mathbf{X}_{\text{curr}\rightarrow\text{short}}]$
\EndIf

\State $\mathbf{X}'_\text{curr} \gets GridPool(\mathbf{X}_T, \alpha_\text{curr})$ \Comment{\ccomment{New current visual token}}

\If{$T > B+1$} \Comment{\ccomment{Out of short-term tokens buffer}}
    \State $\mathbf{X}_{\text{short}\rightarrow\text{long}} \gets GridPool(\mathbf{X}_\text{short}[0], \frac{\alpha_{\text{long}}}{\alpha_{\text{short}}})$
    \State $\mathbf{X}'_{\text{short}} \gets \mathbf{X}_{\text{short}}[1:]$
    \State $s \gets \cos(\mathbf{X}_{\text{long}}[-1], \mathbf{X}_{\text{short}\rightarrow\text{long}})$
    \If{$T>B+2\ \text{and}\ s>\tau$} \Comment{\ccomment{Fuse long-term tokens}}
        \State $\mathbf{X}_{\text{last\_long}} \gets \frac{1}{K+1}(K\mathbf{X}_{\text{long}}[-1]+\mathbf{X}_{\text{short}\rightarrow\text{long}})$
        \State $\mathbf{X}'_{\text{long}} \gets \mathbf{X}_{\text{long}}[:-1] + [\mathbf{X}_{\text{last\_long}}]$
        \State $K'\gets K+1$
    \Else \Comment{\ccomment{Add new long-term token}}
        \State $\mathbf{X}'_{\text{long}} \gets \mathbf{X}_{\text{long}} + [\mathbf{X}_{\text{short}\rightarrow\text{long}}]$
        \State $K'\gets 1$
    \EndIf
\EndIf

\end{algorithmic}
\end{algorithm}

Compared to existing video-based large language models~\cite{zhang2024navid, song2024moviechat, li2023llama}, this online merging strategy significantly reduces inference time, achieving an average of 0.2 seconds per inference. This improvement becomes increasingly notable when handling longer video sequences. A detailed analysis of time efficiency is provided in the Supplementary Materials.



\subsection{Action Planning}
\label{sec:action_planning}

After obtaining the merged visual tokens from semantic features ~\cite{sun2023eva}, we adopt established practices in Vision-and-Language models ~\cite{liu2023llava, li2023llama} to perform vision-language alignment, enabling the large language model (LLM) to effectively interpret visual information.
Specifically, we leverage a cross-modality projector $P_V(\cdot)$ to project all merged visual tokens $X_\text{merged} = \{\mathbf{X}_\text{long}, \mathbf{X}_\text{short}, \mathbf{X}_\text{curr}  \}$ into visual observation tokens that are compatible with the LLM's input representation space:

\begin{equation}
    \label{equ:project}
    \mathbf{E}^V_T = P_V(\mathbf{X}_\text{merged}),
\end{equation}

where the $P_V(\cdot)$ is implemented as a two-layter MLP~\cite{liu2023llava} and optimized in an end-to-end training manner. For instruction encoding, we use the off-the-shelf language tokenizer and embeing layer of LLM (Vicuna-7B ~\cite{chiang2023vicuna}) to encode navigation instruction into language observation tokens $\mathbf{E}^L_{T}$. Then we concatenate the visual observation tokens $\mathbf{E}^V_{T}$, a navigation task indicator $\langle \textit{NAV} \rangle$ and language observation tokens  $\mathbf{E}^V_{T}$  form the final input token sequence. Here, the navigation task indicator $\langle \textit{NAV} \rangle$ is adopted by following ~\cite{zhang2024navid, openai2023gpt4} for accelerating the specific task learning and obtaining consistent output format.  Finally, the complete input token sequence is fed into the LLM to infer four action tokens $\{\mathbf{E}^A_T, \cdots, \mathbf{E}^A_{T+3}\}$, as described below. We include a discussion on the input token format in the Supplementary Material

\begin{tcolorbox}
\textbf{Input:} $\{Long\_term\_tokens\}\{Shot\_term\_tokens\}\\ \{Current\_tokens\} <\!NAV\!> \{Instruction\}$\\
\textbf{Output:} $<\!Action\_0\!>\!<\!Action\_1\!>\!<\!Action\_2\!>\!\\\!<\!Action\_3\!>$
\end{tcolorbox}

The action tokens belong to the discrete action set $\{\texttt{FORWARD}, \texttt{TURN-LEFT}, \texttt{TURN-RIGHT}, \texttt{STOP}\}$. 
Following the standard configuration in existing navigation settings~\cite{savva2019habitat,zeng2024poliformer}, the forward action corresponds to a movement of $25$ cm, and the turning actions represent a $30^\circ$ rotation. This configuration is consistent with all training navigation data (Sec.~\ref{sec:train}). Empirically, we find that predicting the next four steps yields optimal performance, which encourages \name~to forecast long-horizon action sequences while still considering sufficient observations for accurate prediction. This multi-step prediction also supports asynchronous deployment, enabling non-blocking navigation performance in the real world. Please see the Supplementary Material for detailed elaboration.


\section{Data Collection and Training}
\label{sec:train}



To train ~\name for mastering multi-navigation tasks, it is crucial to gather extensive and diverse navigation data across various tasks and environments. However, directly collecting large amounts of real-world navigation data can be prohibitively expensive.
To address this challenge, we propose two key strategies for training \name: First, we collect multi-task navigation data from a wide range of synthetic environments (totaling 861 scenes) using a uniform input and output format, enabling \name to acquire general navigation skills. Second, we co-tune \name with real-world video-based question-answering data, enhancing its ability to interpret real-world images and supporting its open-vocabulary knowledge acquisition.

\begin{figure}[t]
\setlength{\abovecaptionskip}{1pt}
\setlength{\belowcaptionskip}{0pt}
\begin{center}
  \includegraphics[width=1\linewidth]{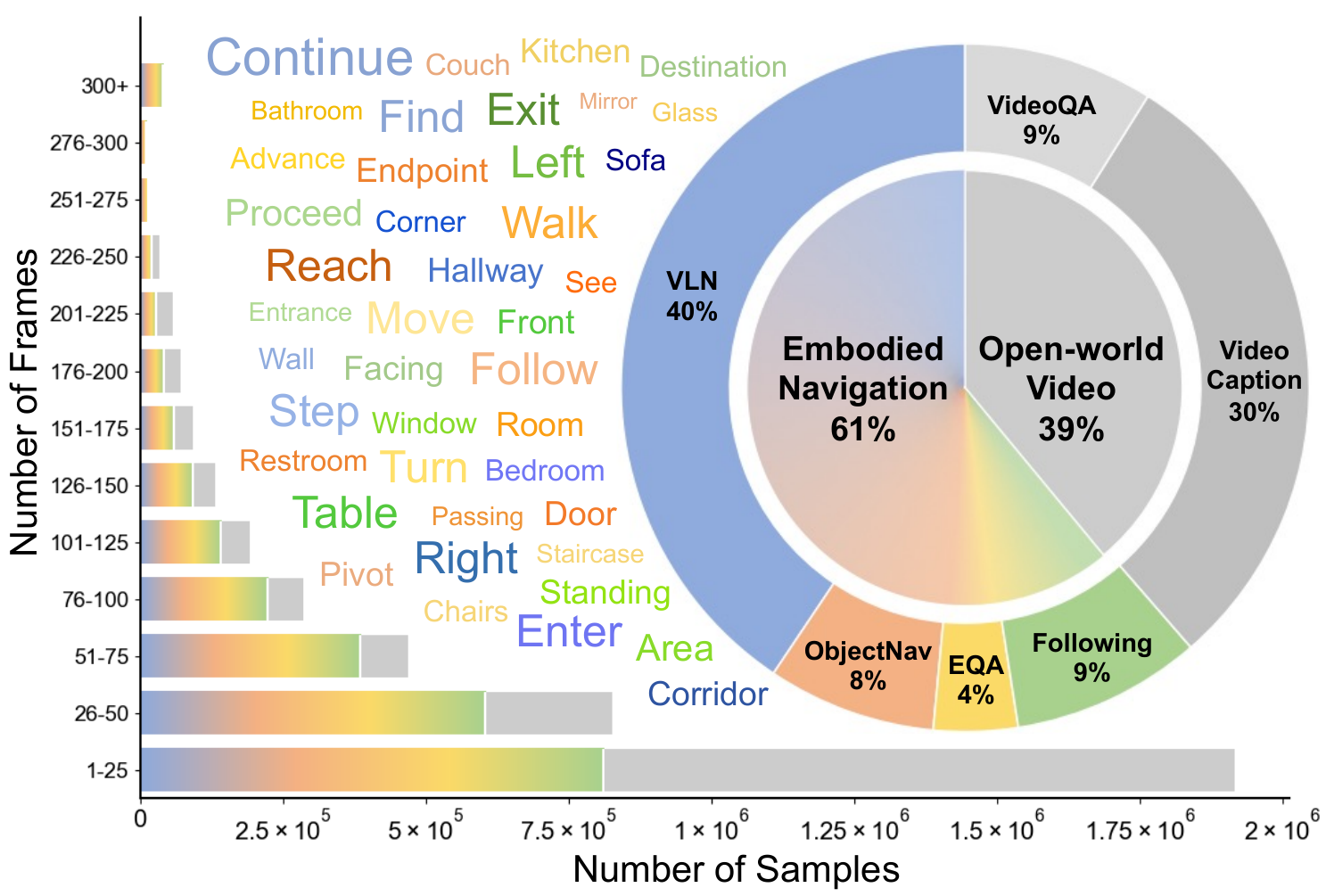}
\end{center}
   \caption{\textbf{Visualization of training data.} We visualize the combination of training data (5.9M), video frame counts, and the most common words in navigation instructions.}
\label{fig:data}
\vspace{-3mm}
\end{figure}


\subsection{Multi-Task Navigation Data.}

\label{sec:data_collection}


We collect the largest multi-task navigation dataset to date within the Habitat simulator environment~\cite{savva2019habitat}, comprising 3.6 million samples across four distinct navigation tasks, as described below. All tasks are curated within a unified framework. A detailed data collection strategy is provided in the Supplementary Materials.


\textbf{\textit{(A) Vision-and-language navigation}}~\cite{Krantz2020BeyondTN, anderson2020rxr} require the agent to interpret and ground instructions in visual observations, effectively combining linguistic and visual information to make sequential decisions. Specifically, the agent has to navigate based on landmarks and motions described in the text and stop nearby the correct destination. Here, we collect 2.4M navigation samples of mainstream VLN datasets, VLN-CE R2R~\cite{Krantz2020BeyondTN} and RxR~\cite{ku2020room}, that focus on continuous environments.


\textbf{\textit{(B) Object Goal Navigation}}~\cite{savva2019habitat} involves an agent navigating an environment to locate a specific object based on provided visual or linguistic cues. This task evaluates the agent's ability to perceive objects, understand scene layout, and execute efficient search strategies. We collected 483k samples from datasets in the Habitat Matterport 3D dataset (HM3D ObjectNav)~\cite{ramakrishnan2021hm3d}. Note that, in HM3D ObjectNav, the agent is required to locate objects from a predefined category set (e.g., \textit{sofa}, \textit{chair}, and \textit{bed}). Nevertheless, experiments demonstrate that our method generalizes to SOTA-level open-vocabulary object goal searching, as shown in Table~\ref{tab:openvocab-objnav}.

\textbf{\textit{(C) Embodied question answering}}~\cite{wijmans2019embodied} requires the agent to navigate to the related area for question answering. It involves spatial reasoning, object description, and understanding contextual information, requiring the ability to integrate perception, language comprehension, and decision-making. Following the setup in main stream EQA methods~\cite{das2018embodied, wijmans2019embodied}, the agent first navigates to the target related to the question, issues a stop action, and then provides an answer. We collect 240k video-action samples and 10k video-answering samples on the MP3D-EQA dataset~\cite{das2018embodied} on Matterport 3D environments~\cite{chang2017matterport3d}.

\textbf{\textit{(D) Human following}}~\cite{islam2019person, francis2023principles} requires the agent to track and follow a human target with a specific description in dynamic and crowded environments, \egno, \textit{``Follow the man in the blue t-shirt.''}. The agent must recognize the appearance of the human, follow the correct person described in the instructions, predict their movement trajectory, and keep an appropriate distance while avoiding obstacles.

\begin{figure}[t]
\begin{center}
  \includegraphics[width=1\linewidth]{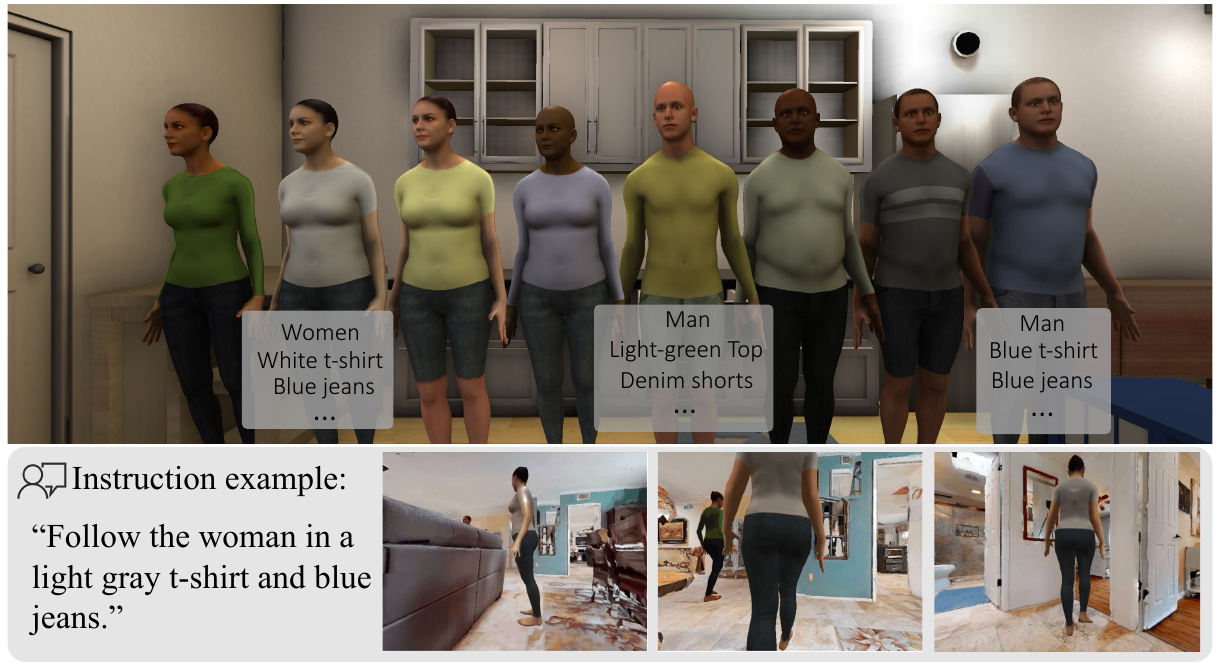}
\end{center}
   \caption{\textbf{Language-described human following benchmark.} We construct our human-following benchmark based on Habitat 3.0~\cite{puig2023habitat} by incorporating textual descriptions for each avatar (eight in total, top row). The robot is required to comprehend these descriptions and accurately follow the designated individual in crowded environments.}
\label{fig:following}
\end{figure}

However, there is currently no human-following dataset that supports language-described human following in crowded environments (multi-person scenarios). To this end, we extend the Habitat 3.0 social navigation benchmark~\cite{puig2023habitat} by (1) adding textual descriptions for each avatar (8 in total, as illustrated in Fig.~\ref{fig:following}), (2) introducing additional distracting human avatars to simulate challenging real-world environments, and (3) deploying the robot and humans in the Habitat Matterport 3D dataset~\cite{yokoyama2024hm3d}, which offers photo-realistic rendering quality and diverse large-scale scenes. The robot and target human are initialized nearby (using the same setting as ~\cite{puig2023habitat}), with randomly moving distracting human avatars. Based on this setup, we collected 544k human-following navigation samples. We also add a detailed description in Supplementary Material. \textit{This benchmark will also be released to benefit the navigation community.}



\textbf{Unified navigation samples.} The data statistics are presented in Figure~\ref{fig:data}. It is worth noting that the number of samples in VLN is relatively larger compared to other tasks. This is because VLN~\cite{Krantz2020BeyondTN, anderson2020rxr} requires the agent to navigate all landmarks described in the instructions, which often results in longer trajectories and, consequently more video-action samples. Here, we collect all navigation samples in a uniform format, including an egocentric RGB video, a natural language instruction, and four corresponding future actions.  All data were collected from synthetic scenes across the Habitat-Matterport 3D (HM3D) and Matterport 3D (MP3D) datasets. We use the default settings of each environment, with a height range of $0.88$ m to $1.25$ m and a robot radius between $0.1$~m and $0.6$ m. This approach helps prevent overfitting to a specific robot embodiment. This approach helps prevent overfitting to a specific robot embodiment. Note that while there exist insightful techniques~\cite{eftekhar2024one, hirose2023exaug} investigating navigation for robots of general sizes, our focus is primarily on uniform multi-task navigation.


\subsection{Training Strategy of Uni-NaVid}

\textbf{Joint training on synthetic and real-world data.} Although we collect navigation data from various environments, the diversity in both observations and instructions remains limited to a specific set of synthetic environments. To incorporate open-world knowledge, we follow previous Vision-and-Language Action models~\cite{zhang2024navid, brohan2023rt}, integrating open-world video question-answering during training. Specifically, we adopt a two-stage training process (a common strategy in Vision-and-Language models~\cite{liu2023llava,li2023llama,song2024moviechat}): (1) First, we exclusively train the cross-modality projector (Equ.~\ref{equ:project}) using the same modality alignment dataset as LLaMA-VID~\cite{li2023llama}. (2) Second, we fine-tune both the projector and the Large Language Model (LLM) using 2.3M video question-answering data from publicly available datasets~\cite{azuma2022scanqa,chen2024panda,li2023llama}, along with 3.6M multi-task navigation samples. During training, we apply the online token merging to both the VQA samples and navigation samples, the only difference is the VAQ samples do not include navigation task indicator  $\langle \textit{NAV} \rangle$.


\textbf{Training configuration.} \name~is trained on a cluster server with 40 NVIDIA H800 GPUs for approximately 35 hours, totaling 1400 GPU hours. For video data, we sample frames at 1 FPS to remove redundant information between consecutive frames. During training, the vision encoder (EVA-CLIP~\cite{sun2023eva}) and large language model (Vicuna-7B~\cite{chiang2023vicuna}) are pre-loaded with default pre-trained weight. Following the training strategy of VLM \cite{liu2023llava}, we optimize the trainable parameters for only 1 epoch.



\section{Experiment}

We conduct experiments to evaluate \name~on three specific aspects: (1) How does \name~perform on individual tasks? (2) Does learning multiple navigation tasks lead to synergistic improvements? (3) Is the key design of our method effective? To evaluate the general-purpose navigation method, we conduct extensive experiments on individual navigation tasks, employing corresponding strong baselines. Additional details are provided in the supplemental material.

\textbf{Benchmarks}. We evaluate our method on various benchmarks across different navigation tasks. Given the diversity of benchmarks spanning various environments and simulators, we meticulously verify the scene splits to ensure no overlap exists between the training and validation scenes across benchmarks.

\begin{itemize} 
\item \textbf{\textit{Vision-and-language navigation}}: We test our method on the validation splits of the VLN-CE R2R~\cite{Krantz2020BeyondTN} and RxR~\cite{anderson2020rxr} benchmarks. 
\item \textbf{\textit{Object goal navigation}}: We use the validation split of the Habitat Matterport 3D (HM3D) dataset~\cite{ramakrishnan2021hm3d}, which requires the agent to find target objects from six categories (sofa, chair, TV, bed, toilet, and plant) in unseen environments. Moreover, to test generalizability, we also evaluate our method on the HM3D-OVON dataset~\cite{yokoyama2024hm3d}, an open-vocabulary object navigation benchmark, in a zero-shot manner. 
\item \textbf{\textit{Embodied question-answering}}: We use the validation split of the MP3D-EQA benchmark~\cite{wijmans2019embodied}. Additionally, we conduct experiments on the more recent Embodied Video Question Answering benchmark, OpenEQA~\cite{majumdar2024openeqa}. 
\item \textbf{\textit{Human following}}: We evaluate our method alongside mainstream approaches on our proposed language-described human following benchmark. 
\item \textbf{\textit{Video understanding}}: We follow the evaluation procedures of existing VQA methods~\cite{li2023llama}. We choose the ScanQA~\cite{azuma2022scanqa},~MSVD~\cite{chen2011collecting},~MSRVTT~\cite{xu2016msr}, and ActivityNet~\cite{caba2015activitynet} datasets. 
\end{itemize}



\textbf{Metrics.} To evaluate navigation performance, we follow the standard evaluation metrics~\cite{anderson2018vision}, including success rate (SR), oracle success rate (OS), success weighted by path length (SPL)~\cite{anderson2018spl}, trajectory length (TL), following rate (FR)~\cite{puig2023habitat}, collision rate (CR)~\cite{puig2023habitat} and navigation error from goal (NE). Note that the success criteria change among different navigation tasks, we therefore use the default success criteria of each benchmark. 
For video understanding evaluation,  we employ widely used metrics following existing works~\cite{azuma2022scanqa, li2023llama}.

\subsection{Deployment Details of Uni-Navid.}

\label{sec:imp_details}

\textbf{Benchmark evaluation.} 
For each navigation task, we adhere to the default settings of each navigation task~\cite{Krantz2020BeyondTN,savva2019habitat,das2018embodied,islam2019person}. All tasks take an online captured RGB video (capturing one frame after each action) and a textual instruction as inputs, and output the next four actions (Sec.~\ref{sec:action_planning}). The robot then executes the predicted actions and calls \texttt{STOP} once the first predicted action is a stop action. For VLN and EQA tasks, we directly use the text instruction provided by the benchmark episodes. For human following and object goal navigation, we transform the target information into an instruction by adding prefixes such as "Search for" or "Follow." Further details can be found in the supplemental material.

It is worth noting that for EQA~\cite{das2018embodied} task, the agent executes navigation actions until a stop command is issued. We then remove the navigation-specific token \texttt{<NAV>} and query the questions using the navigation history. This strategy alleviates the ambiguity for the LLM in deciding whether to navigate or answer a question (See Table~\ref{tab:ablation-study}).

 \textbf{Real-world deployment.} For real-world deployment, we utilize a remote server with an NVIDIA A100 GPU to run \name, which processes observations (along with text instructions) and sends commands to a local robot to execute the predicted actions. \name~requires approximately 0.2 seconds to generate the next four actions. During navigation, the robot asynchronously compresses and uploads the latest observations to the model while executing pending actions. Refer to the supplementary video for real-world navigation performance.

\subsection{Individual Task Results}

\begin{table}[t]
\centering

\resizebox{\linewidth}{!}{
\scalebox{1}{
\setlength{\tabcolsep}{0.7mm}{
\begin{tabular}{l|cccc|ccccc}
\hline
\multirow{2}{*}{Method} & \multicolumn{4}{c|}{Observation} & \multicolumn{5}{c}{VLN-CE R2R Val-Unseen}                                                             \\
                        & Pan.   & Odom.   & Depth   & S.RGB  & TL & \textbf{NE}$\downarrow$ & \textbf{OS}$\uparrow$ & \textbf{SR}$\uparrow$ & \textbf{SPL}$\uparrow$ \\ \hline
HPN+DN$^*$~\cite{krantz2021waypoint}       & \checkmark & \checkmark & \checkmark &            & 7.62  & 6.31          & 40.0          & 36.0          & 34.0          \\
CMA$^*$~\cite{hong2022bridging}          & \checkmark & \checkmark & \checkmark &            & 10.90 & 6.20          & 52.0          & 41.0          & 36.0          \\
\vlnbert$^*\dag$~\cite{hong2022bridging}        & \checkmark & \checkmark & \checkmark &            & 12.23 & 5.74          & 53.0          & 44.0          & 39.0          \\
Sim2Sim$^*$~\cite{krantz2022sim}      & \checkmark & \checkmark & \checkmark &            & 10.69 & 6.07          & 52.0          & 43.0          & 36.0          \\
GridMM$^*$~\cite{wang2023gridmm}       & \checkmark & \checkmark & \checkmark &            & 13.36 & 5.11          & 61.0          & 49.0          & 41.0          \\
HAMT$^*$$\ddag$~\cite{wang2023scaling} & \checkmark & \checkmark & \checkmark &            & –     & 4.80          & –             & 55.0          & 51.0          \\
ETPNav$^*$~\cite{an2023etpnav}       & \checkmark & \checkmark & \checkmark &            & 11.99 & 4.71          & 65.0          & 57.0          & 49.0          \\ \hline
InstructNav~\cite{long2024instructnav} & \checkmark & \checkmark & \checkmark & \checkmark & 7.74 & 6.89 & - & 31.0 & 24.0 \\
AG-CMTP~\cite{chen2021topological}          & \checkmark & \checkmark & \checkmark &            & –     & 7.90          & 39.2          & 23.1          & 19.1          \\
R2R-CMTP~\cite{chen2021topological}          & \checkmark & \checkmark & \checkmark &            & –     & 7.90          & 38.0          & 26.4          & 22.7          \\
LAW~\cite{raychaudhuri2021law}              &            & \checkmark & \checkmark & \checkmark & 8.89  & 6.83          & 44.0          & 35.0          & 31.0          \\
CM2~\cite{georgakis2022cross}              &            & \checkmark & \checkmark & \checkmark & 11.54 & 7.02          & 41.5          & 34.3          & 27.6          \\
WS-MGMap~\cite{chen2022weakly}         &            & \checkmark & \checkmark & \checkmark & 10.00 & 6.28          & 47.6          & 38.9          & 34.3          \\
ETPNav.FF~\cite{wang2024sim}        &            & \checkmark & \checkmark & \checkmark & -     & 5.95          & \textbf{55.8} & 44.9          & 30.4          \\
Seq2Seq~\cite{Krantz2020BeyondTN}          &            &            & \checkmark & \checkmark & 9.30  & 7.77          & 37.0          & 25.0          & 22.0          \\
CMA~\cite{Krantz2020BeyondTN}              &            &            & \checkmark & \checkmark & 8.64  & 7.37          & 40.0          & 32.0          & 30.0          \\
NaVid~\cite{zhang2024navid}            &            &            &            & \checkmark & 7.63  & 5.47          & 49.1          & 37.4          & 35.9          \\
\textbf{\name}       &            &            &            & \checkmark & 9.71  & \textbf{5.58} & 53.3          & \textbf{47.0} & \textbf{42.7} \\ \hline
\end{tabular}
}
}
}
\caption{\textbf{Vision-and-language navigation (R2R).} Comparison on VLN-CE R2R~\cite{Krantz2020BeyondTN} Val-Unseen. $^*$: Methods use high-level action space. $\dag$: Methods use the same waypoint predictor proposed in~\cite{hong2022bridging}. $\ddag$: Methods use additional visual data than MP3D scenes~\cite{chang2017matterport3d}.}
\label{tab:comp-vlnce-r2r}
\end{table} 


\begin{table}[!t]
\centering

\resizebox{\linewidth}{!}{
\scalebox{1}{
\setlength{\tabcolsep}{0.8mm}{
\begin{tabular}{l|ccc|ccccc}
\hline
\multirow{2}{*}{Method}         & \multicolumn{3}{c|}{Observation}     & \multicolumn{5}{c}{VLN-CE RxR Val-Unseen}                             \\ 
 & Odom. & Depth & S.RGB & TL & \textbf{NE}$\downarrow$ & \textbf{OS}$\uparrow$ & \textbf{SR}$\uparrow$ & \textbf{SPL}$\uparrow$ \\ \hline
LAW*~\cite{raychaudhuri2021law}  & \checkmark & \checkmark & \checkmark & 4.01  & 10.87         & 21.0          & 8.0           & 8.0           \\
CM2*~\cite{georgakis2022cross}   & \checkmark & \checkmark & \checkmark & 12.29 & 8.98          & 25.3          & 14.4          & 9.2           \\
WS-MGMap*~\cite{chen2022weakly}  & \checkmark & \checkmark & \checkmark & 10.80 & 9.83          & 29.8          & 15.0          & 12.1          \\
ETPNav.FF~\cite{wang2024sim}                       & \checkmark & \checkmark & \checkmark & -     & 8.79          & 36.7          & 25.5          & 18.1          \\
Seq2Seq*~\cite{Krantz2020BeyondTN} &            & \checkmark & \checkmark & 1.16  & 11.8          & 5.02          & 3.51          & 3.43          \\
CMA*~\cite{Krantz2020BeyondTN}     &            & \checkmark & \checkmark & 5.09  & 11.7          & 10.7          & 4.41          & 2.47          \\
$A^2$Nav$^\dagger$~\cite{chen20232}       &            &            & \checkmark & --    & --            & --            & 16.8          & 6.3           \\
NaVid*~\cite{zhang2024navid}                          &            &            & \checkmark & 10.59 & 8.41          & 34.5          & 23.8          & 21.2          \\
\textbf{Uni-NaVid}                       &            &            & \checkmark & 15.8  & \textbf{6.24} & \textbf{55.5} & \textbf{48.7} & \textbf{40.9} \\ \hline
\end{tabular}
}
}
}
\caption{\textbf{Vision-and-language navigation (RxR).} Comparison on VLN-CE RxR~\cite{ku2020room} Val-Unseen. $^*$: only trained on VLN-CE R2R.}

\label{tab:comp-vlnce-rxr}
\end{table}


\textbf{Comparison on vision-and-language navigation.} We evaluate our method with mainstream baselines on two publicly available benchmarks: VLN-CE R2R~\cite{Krantz2020BeyondTN} and RxR~\cite{ku2020room}. The results are shown in Table~\ref{tab:comp-vlnce-r2r} and Table~\ref{tab:comp-vlnce-rxr}. We find that our methods achieve SOTA-level performance on both datasets using only RGB videos as observations. In comparison to NaVid~\cite{zhang2024navid}, which is also a vision language model that is solely trained on VLN data, our approach demonstrates significant improvements, with a $+25.7\%$ increase in Success Rate (SR) on R2R. For zero-shot methods (InstructNav~\cite{long2024instructnav} and A$^2$Nav~\cite{chen20232}) that use ChatGPT with only text inputs for visual language navigation (VLN), these approaches often face challenges in transitioning between text prompts and visual information, resulting in less than satisfactory outcomes. 
Furthermore, it is important to note that the trajectories in RxR are more diverse and involve longer paths with detailed landmark descriptions, making RxR widely regarded as more challenging than R2R. However, our method achieves consistent performance across both R2R and RxR, with slightly better results on RxR (+$3.6$ SR($\%$)), demonstrating its ability to effectively leverage detailed instructions to navigate diverse trajectories. We add experiments of removing RxR samples in Supplemntal Material, where our method still achive STOA performance (+$23.9$ SR($\%$)) against NaVid.

\begin{table}[!t]
\centering

\scalebox{0.8}{
\setlength{\tabcolsep}{2mm}{
\begin{tabular}{l|ccc|cc}
\hline
\multirow{2}{*}{Method} & \multicolumn{3}{c|}{Observation}     & \multicolumn{2}{c}{HM3D ObjectNav}                       \\ 
                        & Odom.      & Depth      & S.RGB      & \textbf{SR}$\uparrow$ & \textbf{SPL}$\uparrow$ \\ \hline
DD-PPO~\cite{wijmans2020ddppo}                  & \checkmark & \checkmark & \checkmark & 27.9                  & 14.2                   \\
Habitat-Web~\cite{ramrakhya2022habitat}             & \checkmark & \checkmark & \checkmark & 57.6                  & 23.8                   \\
InstructNav~\cite{long2024instructnav}  & \checkmark & \checkmark & \checkmark & 58.0 & 20.9 \\
PIRLNav-IL~\cite{ramrakhya2023pirlnav}                 & \checkmark &            & \checkmark & 64.1         & 27.1                   \\
PIRLNav-IL-RL~\cite{ramrakhya2023pirlnav}                 & \checkmark &            & \checkmark & 70.4         & 34.1                   \\
OVRL~\cite{yadav2023offline}                 & \checkmark &            & \checkmark & 62.0                  & 26.8                   \\
OVRL-v2~\cite{yadav2023ovrl}                 & \checkmark &            & \checkmark & 64.7                  & 28.1                   \\
\textbf{\name}               &            &            & \checkmark & \textbf{73.7}                  & \textbf{37.1}          \\ \hline
\end{tabular}
}
}
\caption{\textbf{Object goal navigation.} Comparison on Habitat Matterport 3D~\cite{ramakrishnan2021hm3d} ObjectNav dataset.}
\label{tab:comp-vlnce-objnav}
\end{table} 


\textbf{Comparison on object goal navigation.} We conduct the experiments on HM3D~\cite{ramakrishnan2021hm3d} to compare \name~with mainstream methods~\cite{wijmans2020ddppo,ramrakhya2022habitat,ramrakhya2023pirlnav,yadav2023offline,yadav2023ovrl} that also learn from ObjectNav data. The results, shown in Table~\ref{tab:comp-vlnce-objnav}, demonstrate that our approach achieves the best performance. Note that methods not utilizing odometry face challenges as they must rely on implicit memory to retain the historical trajectory. Nevertheless, \name~still achieves significant gains in SR (+$4.7\%$) and SPL (+$8.8\%$) compared to previous state-of-the-art methods. Additionally, we believe our method’s ObjectNav performance can be further enhanced by incorporating reinforcement learning techniques, as demonstrated by PIRLNav~\cite{ramrakhya2023pirlnav} and Poliformer~\cite{zeng2024poliformer}.

\begin{table}[!t]
\centering

\scalebox{0.8}{
\setlength{\tabcolsep}{2mm}{
\begin{tabular}{l|cc|cc|cc}
\hline
\multirow{2}{*}{Method} & \multicolumn{2}{c|}{VAL SEEN}   & \multicolumn{2}{c|}{\makecell{VAL SEEN \\ SYNONYMS}} & \multicolumn{2}{c}{VAL UNSEEN} \\
& \textbf{SR}$\uparrow$ & \textbf{SPL}$\uparrow$ & \textbf{SR}$\uparrow$ & \textbf{SPL}$\uparrow$ & \textbf{SR}$\uparrow$ & \textbf{SPL}$\uparrow$ \\ \hline

BC & 11.1 & 4.5 & 9.9 & 3.8 & 5.4 & 1.9 \\
DAgger & 11.1 & 4.5 & 9.9 & 3.8 & 5.4 & 1.9 \\
RL & 18.1 & 9.4 & 15.0 & 7.4 & 10.2 & 4.7 \\
BCRL & 39.2 & 18.7 & 27.8 & 11.7 & 18.6 & 7.5 \\
DAgRL & \textbf{41.3} & \textbf{21.2} & 29.4 & 14.4 & 18.3 & 7.9 \\
VLFM$^*$~\cite{yokoyama2024vlfm} & 35.2 & 18.6 & 32.4& 17.3 & 35.2 & 19.6 \\
DAgRL+OD~\cite{yokoyama2024hm3d} & 38.5 & 21.1 & 39.0 & 21.4 & 37.1 & \textbf{19.8} \\
\textbf{\name}$^*$ & \textbf{41.3} & 21.1 & \textbf{43.9} & \textbf{21.8} & \textbf{39.5} & \textbf{19.8} \\ \hline
\end{tabular}
}
}
\caption{\textbf{Object goal navigation.} Comparison on HM3D-OVON~\cite{yokoyama2024hm3d}. $^*:$ denotes zero-shot methods.}
\label{tab:openvocab-objnav}
\end{table} 


To evaluate the generalization ability for open-vocabulary objects, we evaluate our method on the open-vocabulary object goal navigation benchmark (HM3D-OVON~\cite{yokoyama2024hm3d}) in a zero-shot manner. The results in Table~\ref{tab:openvocab-objnav} demonstrate that our method achieves significant improvement over the zero-shot method (VLFM~\cite{yokoyama2024vlfm}) and even outperforms the fine-tuned method (DAgRL+OD~\cite{yokoyama2024hm3d}) on the VAL SEEN and VAL UNSEEN splits. This proves the generalizability of our method.


\begin{table}[!t]
\centering

\scalebox{0.9}{
\setlength{\tabcolsep}{0.8mm}{
\begin{tabular}{l|ccc|c}
\hline
\multirow{2}{*}{Method} & \multicolumn{3}{c|}{Action Type}     & MP3D EQA               \\
                        & D.E.         & C.E.         & GT         & \textbf{ACC}$\uparrow$ \\ \hline
NaviLLM~\cite{zheng2023towards}                 & \checkmark &            &            & 44.5                   \\
\textbf{\name}               &            & \checkmark &            & \textbf{47.3}                   \\ \hline
EQA(habitat-lab)~\cite{das2018embodied}         &            &            & \checkmark & 46.0                   \\
NaviLLM~\cite{zheng2023towards}                 &            &            & \checkmark & 47.4                   \\
\textbf{\name}               &            &            & \checkmark & \textbf{54.4}                   \\ \hline
\end{tabular}
}
}
\caption{\textbf{Embodied question answering.} Comparison on Habitat Matterport3D EQA dataset~\cite{das2018embodied}.}
\label{tab:comp-vlnce-eqa}
\end{table} 

\textbf{Comparison on embodied question answering.} The evaluation results on MP3D-EQA~\cite{wijmans2019embodied} are presented in Table~\ref{tab:comp-vlnce-eqa}. Despite navigating in continuous environments (CE), our method outperforms existing approaches (\textit{e.g.,} NaviLLM~\cite{zheng2023towards} leverage the same evaluation strategy in Sec.~\ref{sec:imp_details}) that operate within discrete landmark-based environments (DE). Moreover, when provided with the ground truth (GT) navigation trajectory, our method shows a significant improvement, demonstrating its ability to understand navigation history effectively. We also report our performance on the more challenging EM-EQA benchmark, OpenEQA~\cite{majumdar2024openeqa}, in the Supplemental Material, which includes more complex questions. Our method achieves comparable performance to GPT-4V with scene captions~\cite{majumdar2024openeqa}.

\begin{table}[!t]
\centering

\scalebox{0.9}{
\setlength{\tabcolsep}{0.8mm}{
\begin{tabular}{l|cc|ccc}
\hline
\multirow{2}{*}{Method} & \multicolumn{2}{c|}{Observation} & \multicolumn{3}{c}{Human Following Dataset} \\
& H.Det. & S.RGB & \hspace{0.3cm}\textbf{SR}$\uparrow$\hspace{0.2cm} & \hspace{0.2cm}\textbf{FR}$\uparrow$\hspace{0.2cm} & \hspace{0.2cm}\textbf{CR}$\downarrow$ \\
\hline

PoliFormer~\cite{zeng2024poliformer}     &  & $\checkmark$ & 2.79 & 20.35 & 2.93 \\
PoliFormer$^*$~\cite{zeng2024poliformer}     & $\checkmark$ & $\checkmark$ & 14.67 & 37.14 & 4.29 \\
PoliFormer$\dag $~\cite{zeng2024poliformer}     & $\checkmark$ & $\checkmark$ & 25.29 & 47.16 & 6.78 \\
IBVS$^*$~\cite{gupta2016novel} & $\checkmark$ & $\checkmark$ & 46.08 & 62.64 & 0.84 \\
IBVS$\dag $~\cite{gupta2016novel} & $\checkmark$ & $\checkmark$ & 50.58 & 68.89 & \textbf{0.80} \\
\textbf{\name}           & & $\checkmark$ & \textbf{61.21} & \textbf{71.93} & 2.07 \\ \hline
\end{tabular}
}
}
\caption{\textbf{Human following.} Comparison on Human Following Dataset. $^*$: Methods use GroundingDINO~\cite{liu2023grounding} as the open-vocabulary human detector. $\dag$: Methods use the ground-truth bounding box provided by the simulator.}
\label{tab:human-following}
\end{table}

\textbf{Comparison on human following.} We compared our method with two most relative methods  PoliFormer~\cite{zeng2024poliformer} and IBVS~\cite{gupta2016novel}. Since both methods require a specific human bounding box as input, obtained from an upstream algorithm, we use the bounding box from the open-world object detector GroundingDINO~\cite{liu2023grounding} and the ground truth provided by the simulator to evaluate the human following performance of the comparison methods under various setups. As shown in Table~\ref{tab:human-following}, \name~outperforms the comparison methods on both SR (+$21.0\%$) and FR (+$4.4\%$) while maintaining low CR under any setup, even when they use ground truth bounding boxes as input. This demonstrates that \name~can effectively infer instructions and follow the correct human, as well as predict the human’s movement patterns accurately. We include additional human-following experiments in various environments, such as HSSD~\cite{khanna2024habitat} and MP3D~\cite{chang2017matterport3d}, in the Supplemental Material. Our method consistently demonstrates SOTA performance across these settings.

\begin{figure*}[t]
\setlength{\abovecaptionskip}{1pt}
\begin{center}
  \includegraphics[width=1 \linewidth]{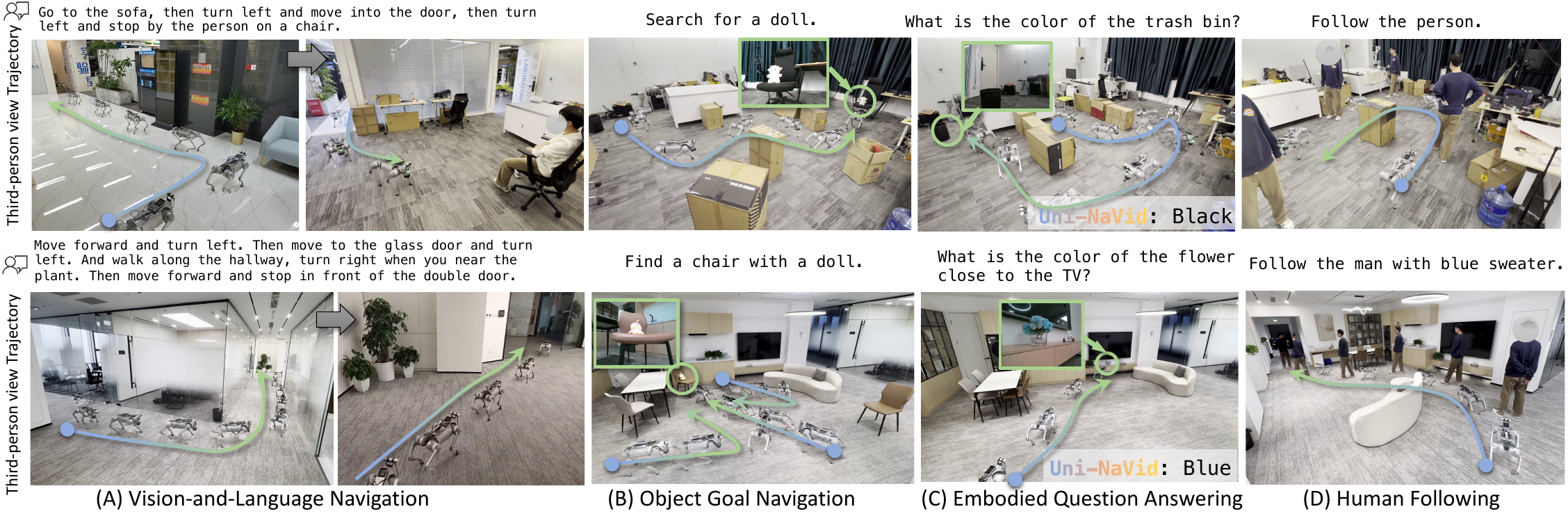}
\end{center}
   \caption{\textbf{Visual results of \name~in real-world.} We deploy \name~across diverse environments to execute instructions in a zero-shot setting. We provide third-person views with robot’s trajectory, showing effective navigation performance. We indicate the starting point as \textcolor{myblue}{a blue dot} and the ending point as \textcolor{mygreen}{a green arrow}. }
\label{fig:gallery}
\end{figure*}


\begin{table}[!t]

\centering

\scalebox{0.78}{
\setlength{\tabcolsep}{0.8mm}{
\begin{tabular}{l|ccccc}
\hline
\multirow{2}{*}{Method} & \multicolumn{5}{c}{ScanQA}                                                                                                 \\
                        & EM $\uparrow$            & BLUE-1  $\uparrow$                 & ROUGE $\uparrow$          & \multicolumn{1}{l}{METEOR} $\uparrow$ & \multicolumn{1}{l}{CIDEr} $\uparrow$ \\ \hline
V.N.+MCAN~\cite{yu2019deep}            & 19.71          & 29.46                     & 30.97          & 12.07                      & 58.23                     \\
S.R.+MCAN~\cite{yu2019deep}          & 20.56          & 27.85                     & 30.68          & 11.97                      & 57.56                     \\
3D-LLM(flamingo)~\cite{hong20233d}        & 23.2           & 32.60                    & 34.80          & 13.5                       & 65.6                      \\
NaviLLM~\cite{zheng2023towards}                 & 26.27          & 39.73                  & 40.23          & 16.56                      & 80.77                     \\
BridgeQA~\cite{mo2024bridging}                & \textbf{31.29} & 34.49          & 43.26          & 16.51                      & 83.75                     \\
\textbf{Uni-NaVid}               & 28.01          & \textbf{46.85}          & \textbf{45.74} & \textbf{19.24}             & \textbf{94.72}            \\ \hline
\end{tabular}
}
}
\caption{\textbf{Embodied video question answering.} Comparison on ScanQA~\cite{azuma2022scanqa} benchmark.}
\label{tab:comp-scanqa}
\end{table}

\begin{table}[!t]
\centering

\scalebox{0.9}{
\setlength{\tabcolsep}{0.8mm}{
\begin{tabular}{l|cc|cc|cc}
\hline
\multirow{2}{*}{Method} & \multicolumn{2}{c|}{MSVD-QA}    & \multicolumn{2}{c|}{MSRVTT-QA}  & \multicolumn{2}{c}{ActivityNet-QA} \\
                        & Acc$\uparrow$ & Score$\uparrow$ & Acc$\uparrow$ & Score$\uparrow$ & Acc$\uparrow$   & Score$\uparrow$  \\ \hline
VideoLLaMA~\cite{zhang2023video}   & 51.6          & 2.5          & 29.6          & 1.8          & 12.4 & 1.1 \\
VideoChat~\cite{li2023videochat}    & 56.3          & 2.8          & 45.0          & 2.5          & 26.5 & 2.2 \\
VideoChatGPT~\cite{maaz2023video} & 64.9          & 3.3          & 49.3          & 2.8          & 35.2 & 2.7 \\
BT-Adapter~\cite{liu2024bt}   & 67.5          & 3.7          & 57.0          & 3.2          & 45.7 & 3.2 \\
Chat-UniVi~\cite{jin2024chat}   & 65.0          & 3.6          & 54.6          & 3.1          & 45.8 & 3.2 \\
LLaMA-VID~\cite{li2023llama}    & 69.7          & 3.7          & 57.7          & 3.2          & 47.4 & 3.3 \\
VideoChat2~\cite{li2024mvbench}   & 70.0          & \textbf{3.9} & 54.1          & 3.3          & 49.1 & 3.3 \\
Video-LLaVA~\cite{lin2023video}  & 70.7          & \textbf{3.9} & 59.2          & \textbf{3.5} & 45.3 & 3.3 \\
ST-LLM~\cite{liu2025st}       & \textbf{74.6} & \textbf{3.9} & \textbf{63.2} & 3.4          & 50.9 & 3.3 \\
\textbf{\name}               & 69.6          & \textbf{3.9}    & 59.3          & \textbf{3.5}    & \textbf{51.4}   & \textbf{3.7}     \\ \hline
\end{tabular}
}
}
\caption{\textbf{Video question answering.} Comparison with leading methods (all based on Vicuna-7B~\cite{chiang2023vicuna}) on VQA benchamarks. }
\label{tab:comp-videoqa}
\end{table} 


\begin{figure}[t]
\begin{center}
  \includegraphics[width=1 \linewidth]{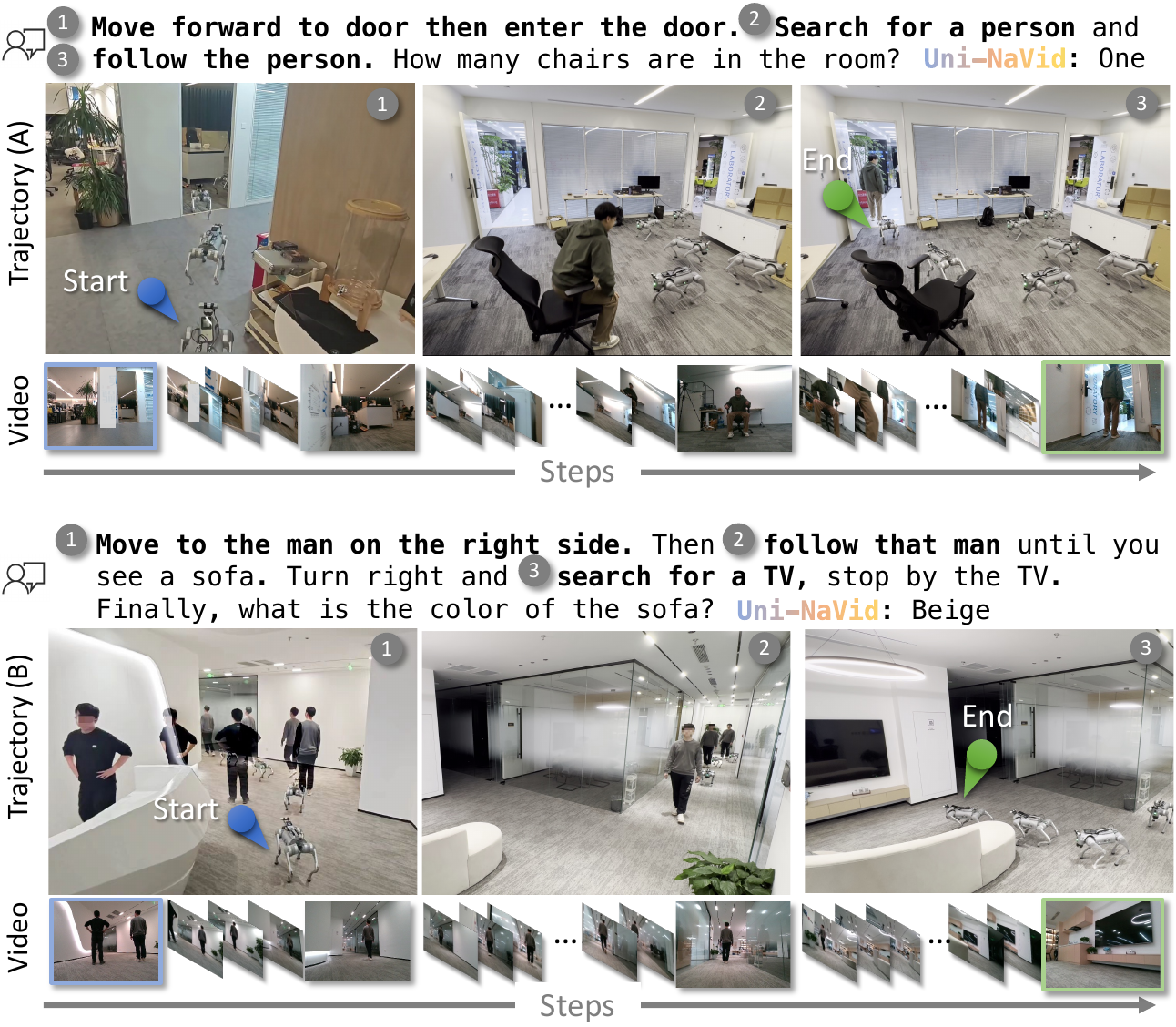}
\end{center}
   \caption{\textbf{Vivusal results of \name{} on compostional tasks.} The agent is required to execute complex instructions involving multiple navigation tasks. Our method successfully accomplishes these navigation tasks sequentially. Notably, both the instructions and environments are novel to our approach. Please refer to the supplementary videos.}
\label{fig:comp_task_vis}
\end{figure}


\textbf{Comparison on video question answering.} We first evaluate our method on ScanQA~\cite{azuma2022scanqa} on Tab.~\ref{tab:comp-scanqa}. Compared to mainstream baselines, we find that \name~archives the best performance on four metrics, including BLUE-1 ($+17.9\%$), ROUGE ($+5.7\%$), METEOR ($+16.2\%$), and CIDEr ($+13.1\%$). This proves the superiority of our methods on spatial scene understanding. Note that the EM metric requires an exact match between the question and answer, which is not well-suited to our method, as it is designed to learn from diverse data and generate flexible responses. 

We further evaluate our method on open-ended video question-answering benchmarks~\cite{chen2011collecting, xu2016msr, caba2015activitynet}, as presented in Table~\ref{tab:comp-videoqa}. To ensure a fair comparison, we focus on methods that employ the same large language model backbone (Vicuna-7B~\cite{chiang2023vicuna}). The results indicate that even after extensive token merging (Sec.~\ref{sec:token_merge}), \name achieves performance comparable to state-of-the-art methods. This demonstrates the effectiveness of both our token merging and training strategies, while also highlighting robust open-world understanding capabilities.

\subsection{Qualitative Results in Real-World}

We conducted extensive experiments on real-world environments (experiment details are provided in the supplemental material) under diverse environments in a zero-shot manner. Notably, both the instructions and environments are novel to our method. We first evaluated the performance of individual navigation tasks (Fig.~\ref{fig:gallery}), including (A) vision-and-language navigation, (B) object goal navigation, (C) embodied question answering, and (D) human following. We found that \name{} can understand diverse instructions and demonstrates impressive performance in long-horizon navigation tasks (e.g., navigating across hallways and entering rooms), as well as in searching for out-of-vision objects and answering subsequent questions. Moreover, the agent is capable of following a human even when the person's appearance deviates from the description of the avatar in the human-following dataset (Sec.~\ref{sec:data_collection}). The statistics of the corresponding real-world experiments can be found in the Supplemental Material.

\begin{figure}[t]
\begin{center}
  \includegraphics[width=1 \linewidth]{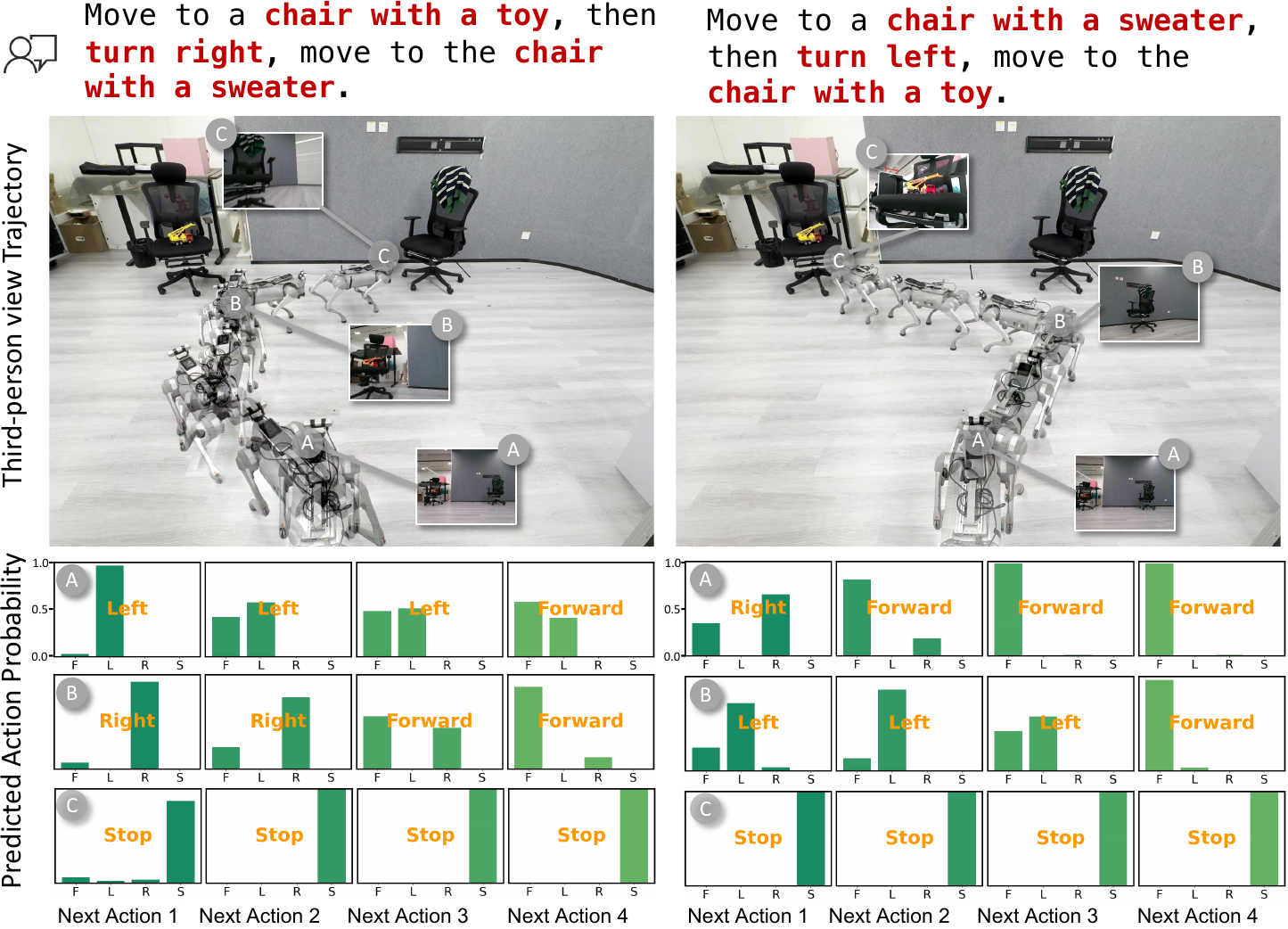}
\end{center}
   \caption{\textbf{Action prediction on the VLN tasks.} We evaluate \name{} on challenging open-vocabulary objects, requiring it to recognize the target objects and follow the specified motions. We provide the predicted action probabilities (for the next four actions) to demonstrate its break-in navigation capability. }
\label{fig:open_vocab_nav}
\end{figure}

In addition to individual navigation tasks, we also evaluate our method on more complex instructions involving multiple navigation tasks (Fig.~\ref{fig:comp_task_vis}). In this scenario, the agent is required to sequentially execute the navigation tasks described in the language instructions. Our model demonstrates impressiove performance in aligning the current navigation process with the instructions to reason about the current state of navigation.
Furthermore, we provide a detailed illustration of action prediction during navigation in Fig.~\ref{fig:open_vocab_nav}, where we plot the predicted action probabilities of \name{}. Notably, with only slight differences in object descriptions, \textit{e.g.,} \textit{'chair with a toy'} and \textit{'chair with a sweater'}. Specifically, our method successfully distinguishes between the locations and predicts actions accordingly. Interestingly, the action probabilities (for the next four actions) reveal a sequential order of actions: first turning right/left, followed by moving forward. We provide additional visual results of our method in the Supplemental Material and encourage the audience to view our video, which showcases the real-world performance of our method.


\begin{figure}[t]
\begin{center}
  \includegraphics[width=1 \linewidth]{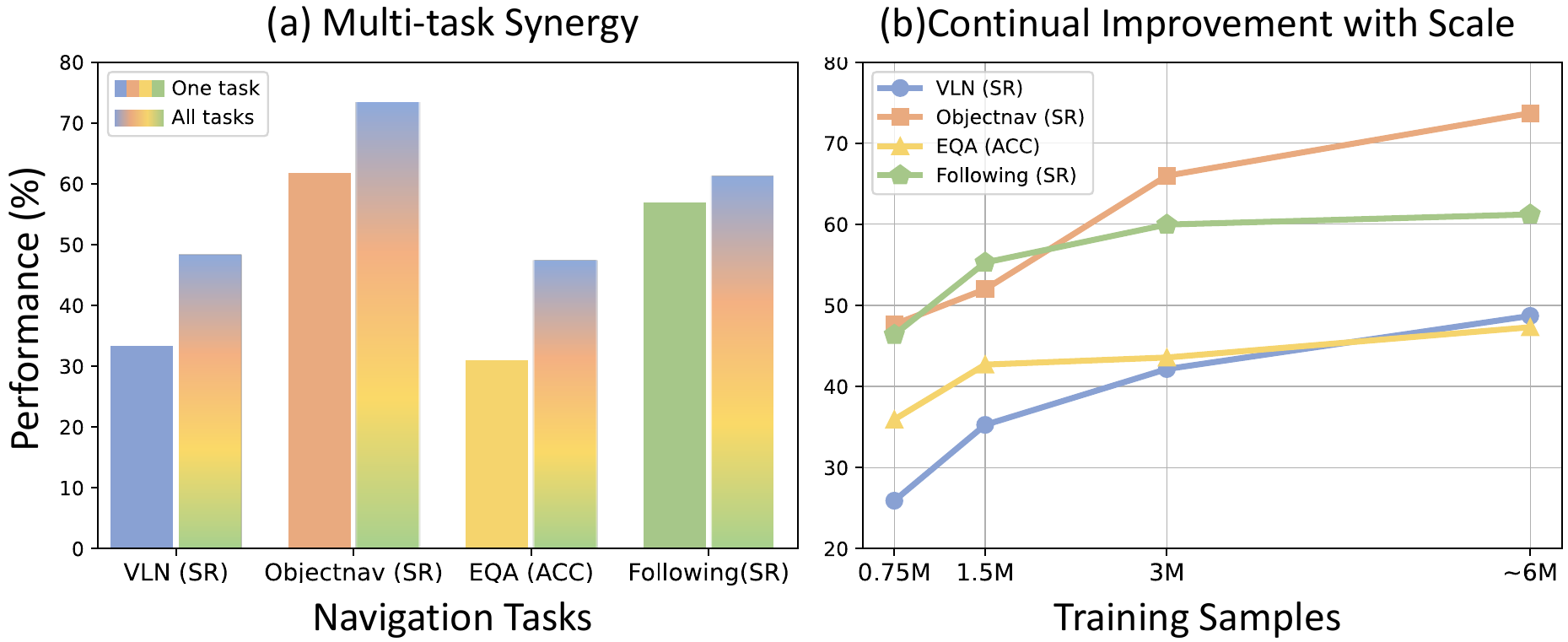}
\end{center}
   \caption{\textbf{Comparsion on multi-task training and data scale.} (a) We present the multi-task synergy of our method, illustrating the performance comparison between training with a single task and training with multiple tasks; (b) we demonstrate the performance across different navigation tasks under varying numbers of training samples.}
\label{fig:visual_results}
\end{figure}


\subsection{Ablation Study}

\textbf{Visualization of training strategy.} We present a visualization of the training strategy’s performance in Figure~\ref{fig:visual_results}. In Fig.~\ref{fig:visual_results} (a), we compare training on a single navigation task with training across multiple tasks. The results demonstrate the synergistic benefits of multi-task learning, which yields consistent performance improvements across all navigation tasks. Notably, VLN, ObjectNav, and EQA exhibit more significant improvements, while Following shows relatively smaller gains. We attribute this difference to the lower reliance of the Following task on historical context. Additionally, we investigate the influence of data scale on navigation performance (Figure~\ref{fig:visual_results} (b)). We observe that performance improves across all navigation tasks with larger data volumes. However, the incremental gain diminishes (from 3M to 6M samples), potentially due to limitations in the data diversity of simulators. Specifically, for the Following task, the reason for the slower convergence is the heavy occlusion caused by obstacles or other humans. This highlights the need for more high-quality following data samples, which can enable our model to learn more effectively and perform better in highly dynamic environments.






\textbf{Ablation on training strategy and architecture.} We conduct experiments to evaluate the effectiveness of the training strategy and token merging designs (Tab.~\ref{tab:ablation-study}). Our results indicate that the absence of \texttt{<NAV>} and VQA data leads to a performance decline across all tasks, similar findings can be found in~\cite{chen2023minigpt, zhang2024navid}. Notably, the performance drop is most obviously in EQA, as the lack of \texttt{<NAV>} special token makes the model misinterpret whether it should answer questions or output actions. Additionally, without VQA data, the agent’s ability to answer questions drops significantly, almost rendering it incapable of correctly answering questions. We believe this is due to the catastrophic forgetting problem in LLMs, where the model loses open-world knowledge by being trained solely on navigation-related data.

From the performance of different memory designs, we find that both short-term and long-term memory visual tokens contribute to performance improvements. In particular, the VLN task shows the most significant performance drop ($-80.3\%$ SR) when visual memory is removed, as the lack of memory hinders the alignment of visual history with instructions. For the Following task, the absence of memory results in only a minor performance decline ($-8\%$ SR), as this task primarily relies on recent frames to track the target. 
Additional ablation studies on architecture and hyperparameters are provided in the Supplementary Material.

\begin{table}[!t]
\centering

\scalebox{0.75}{
\setlength{\tabcolsep}{0.72mm}{
\begin{tabular}{lll|cccc}
\hline
\multicolumn{3}{l|}{Type}             & VLN (SR$\uparrow$) & ObjNav (SR$\uparrow$) & EQA (ACC$\uparrow$) & Follow (SR$\uparrow$) \\ \hline
\multicolumn{3}{l|}{No $<$Nav$>$ token} & 35.2 & 69.1 & 20.4 & 55.1 \\
\multicolumn{3}{l|}{No VQA data}        & 40.5 & 50.6 & 1.19 & 58.8 \\ \hline
\multicolumn{3}{l|}{Curr.}              & 9.61 & 44.3 & 32.5 & 56.3 \\
\multicolumn{3}{l|}{Curr.+Short.}       & 39.7 & 67.8 & 44.1 & 59.7 \\
\multicolumn{3}{l|}{\textbf{Curr.+Short.+Long.}} & \textbf{48.7}      & \textbf{73.7}         & \textbf{47.3}       & \textbf{61.2}         \\ \hline
\end{tabular}
}
}
\caption{\textbf{Ablation study on training strategy and architecture}. For each ablation type, we retrain the entire model and evaluate its performance across four navigation tasks.}
\label{tab:ablation-study}

\end{table}

\section{Limitations}

Despite the promising results, \name{} has several limitations. \textit{\textbf{First}}, \name{} is trained and evaluated on four well-defined navigation tasks, while there exists a large body of literature on insightful and practical navigation datasets~\cite{wang2024find, Zhang2024VisionandLanguageNT}. We believe that collecting data from these datasets could further enhance the navigation capabilities of our method.
\textit{\textbf{Second}}, our method is designed to acquire multi-task navigation capabilities under the assumption that the robot is of standard size (see Section~\ref{sec:data_collection}). To extend it to robots of general sizes, a convincing approach is to incorporate prior knowledge of the robot's size, as demonstrated in~\cite{eftekhar2024one, hirose2023exaug}. \textit{\textbf{Third}}, our method is currently limited to predicting simple trajectories composed of a short horizon of future low-level discrete actions. This limitation could be alleviated by extending the moel to predict continuous and smooth trajectories with techniques from motion planning~\cite{shah2023gnm, sridhar2024nomad} or autonomous driving~\cite{chen2024vadv2, liao2024diffusiondrive}.




\section{Discussion and Conclusion} 
\label{sec:conclusion}


In this paper, we introduce an efficient vision-language-action (VLA) model, \name, designed to acquire general embodied navigation skills through learning multi-task navigation data. To efficiently encode the online-captured video sequences during navigation, we develop an online visual token merging mechanism that separately processes current observations, short-term observations, and long-term observations. This design enables our approach to operate at an average speed of 5 Hz. We also collect 3.6 million navigation data points across four highly demanded embodied navigation tasks, including vision-and-language navigation, object goal navigation, embodied question answering, and human following. Extensive experiments and ablation studies demonstrate that our method achieves SOTA-level performance using only monocular videos as input, highlighting our model's superior capability in learning multiple navigation tasks. Moreover, we deploy \name{} in real-world environments, demonstrating impressive generalizability and versatile navigation performance in real worlds.

\textbf{Future works.} Our work serves merely as a starting point of general-purpose navigation, and we hope it will inspire future directions in this field:

\begin{itemize}
    \item \textit{\textbf{Benchmarking.}} With the consistent development of embodied navigation, there is a growing need for general-purpose navigation benchmarking. Such a benchmark would help researchers better position their work and drive progress in the navigation community.
    \item \textit{\textbf{Architecture.}} We would like to further enhance the practicality of our architecture by tackling very long-horizon tasks (e.g., navigating across buildings) and incorporating advanced motion planning techniques~\cite{sridhar2024nomad, chen2024vadv2}.
    \item \textit{\textbf{Application.}} We would like to apply our method to applications such as robotic guide dogs and home service robots. Additionally, we are excited to extend this technique to other embodied AI tasks, such as mobile manipulation \cite{zhang2024gamma, wu2024tidybot++, liu2024ok}.
\end{itemize}



\bibliographystyle{plainnat}
\bibliography{references}

\clearpage
\tableofcontents  

\section{Task Definition}
We introduce the details of four embodied navigation tasks that are included in our paper.

\subsection{Vision-and-language Navigation (VLN)}

Vision-and-language navigation~\cite{Krantz2020BeyondTN} requires the agent to follow the instruction by moving between given landmarks and stopping at the described destination within unseen environments. The instruction of VLN is at free-form which describes a trajectory of landmarks and the motions between these landmarks. The landmarks and motions are open-vocabulary. VLN is widely regarded as a challenging task because it has to understand the free-form instruction, align the navigation history with instruction, and do path planning. Despite the fact that many works consider using pre-build landmark graphs~\cite{zhou2023navgpt,long2024instructnav,zhou2025navgpt} to simplify VLN, we consider a more practical setting that uses a continuous environment (VLN-CE~\cite{Krantz2020BeyondTN}). Currently, there are two mainstream VLN-CE datasets: VLN-CE R2R~\cite{Krantz2020BeyondTN} and VLN-CE RxR~\cite{ku2020room}. We provide examples of these instructions:

\begin{itemize}
    \item \textbf{VLN-CE R2R}: Walk down the hallway along the banister railing on the upper floor of the home. Walk through the open door next to the staircase. Walk into the room, which has a couch and chairs around a coffee table.
    \item \textbf{VLN-CE RxR}: You are starting in the corner of a living room. Turn around to find a clock hanging on the wall in the hallway. Take two steps toward it. Turn right and walk straight, passing between the blue couch and the kitchen. You should now be looking through windows into the backyard. To your right is an open patio, and to your left are four framed black-and-white pictures. You’ve completed the instructions.
\end{itemize}

In the benchmark VLN-CE R2R and VLN-CE RxR, the agent is required to navigate each of the landmarks in the given order of the instruction and stop within 3 meters of the destination. The max navigation step is 500 steps.

\subsection{Object Goal Navigation (ObjectNav)}

In ObjectNav~\cite{savva2019habitat}, an agent is initialized at a random starting position and orientation in an unseen environment and asked to find an instance of an object category (‘find a chair’) by navigating to it. 
The agent should explore the environments to track the location location that could locate target objects, and then identify the target object and stop nearby. Specifically,  we follow the Habitat-matterport 3D dataset, which requires the agent to search a specific category object, which could be one of a category set including couch, bed, chair, toilet, plant, and TV. In the benchmark HM3D, the trajectory is considered a success if the target stops within 1 meter of the target object under 500 steps.

\subsection{Embodied Question Answering (EQA)}
Embodied Question Answering~\cite{das2018embodied} is a complicated task, which involves answering questions by interacting with and navigating within an unseen 3D environment. Following the setting of MP3D-EQA, the agent is first given a question related to an object, like the color or location, and then the agent is required to identify the described target and return a natural language that directly answers the question. Here are some examples of the questions:
\begin{itemize}
    \item  What room is the chair located in?
    \item What color is the bed?
    \item What color is the couch in the living room?
\end{itemize}

In the benchmark MP3D-EQA, the agent is required to find the target and answer the question within 500 steps. And the episode will be considered a success if the answer matches the GT answer.

\subsection{Human Following}
The three tasks discussed above primarily address static and fixed scenarios, with limited applicability to dynamic environments. In contrast, human following \cite{gupta2016novel, islam2019person, zu2024language}, as a classic robot target tracking problem, focuses on enabling robots to interact effectively in dynamic settings, making it particularly well-suited for real-world applications.

Traditional human following tasks often rely on modular-based approaches \cite{honig2018toward}, mainly focusing on the robot’s following strategies \cite{van2023human} and controller design \cite{kastner2022human}. These studies typically assume that the target to be tracked is either known or specified at the beginning, an assumption that fails to account for the complexities of real-world environments. In this paper, we introduce a novel human-following task driven by natural language instruction, where the robot should locate and follow a specific human that aligns with the given description in a potentially crowded and dynamic environment.

We developed a benchmark for this task based on Habitat 3.0 \cite{puig2023habitat}. In each episode, multiple humans with diverse characteristics are randomly placed in the scene, and the robot is initialized near the target human to be followed, ensuring that this human is within the robot’s initial observation. The robot interprets natural language instructions, such as \textit{``follow the man wearing a blue shirt and black pants''} or \textit{``stay behind the woman in yellow''}, to locate the individual that best matches the description. It then tracks and follows the person’s movements, dynamically adapting until reaching the pre-defined navigation goal. An episode is deemed successful if the robot stops within 2 meters of the correct human and faces him/her at the end.

\section{Implementation Details}
In this section, we provide more implementation details of \name.

\begin{figure}
\begin{center}
  \includegraphics[width=1 \linewidth]{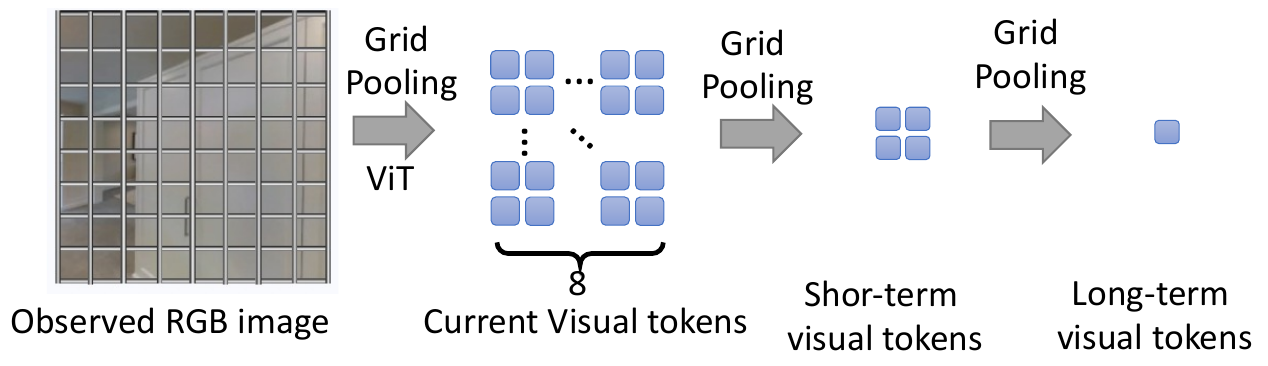}
\end{center}
   \caption{\textbf{Grid pooling}. We add a visualization of grid pooling across different types of observations.}
\label{fig:gird_pooling}
\end{figure}

\subsection{Online Visual Token Merging}
Online visual token merging is the key technique of \name~to enable efficient encoding of a long horizon of ego-centric videos. 
Specifically, we organize the online captured frames. When the model only receives one frame, that frame becomes the current observation. Then we extract the visual tokens and leverage grid pooling (See Fig.~\ref{fig:gird_pooling}). We split the image into $8\times8$ and ach grid is conducted an average operation, leading to the final 64 tokens for current observations. Then with more incoming frames, we perform grid pooling to older current observation tokens, which leads to $2\times2$ tokens, and append them into a short-term memory buffer.

If the time step is over 65 steps, then the oldest short-term frame will be pooped out and then is performed grid pooling to 1 token. This token is a long-term visual token which then will be inserted into the long-term visual token list if the long-term visual token list is empty or the cos similarity is smaller than $\tau$. Here, we use $\tau=0.95$, which is obtained empirically~\cite{song2024moviechat}, which achieves a balance between the efficiency and effective. 

\subsection{Token Organization}

To facilitate the understanding of the large language model, we have to organize the tokens, including observation tokens, instruction tokens, and special tokens. 
Specifically, we use observation indicator tokens to indicate the following parts are visual tokens. Besides, we add an image separator token between the adjacent visual tokens of each frame (following existing VLMs~\cite{Hong2024CogVLM2VL, liu2025st}), this is crucial to distinguish the visual information inherent from different frames. Finally, if the task is navigation, we add a navigation special token \texttt{<NAV>} to indicate the task is navigation. Once the model understands the navigation special token, it will directly output action tokens. 

Note that, it is important to use a navigation special token which could address the ambiguity problem under the embodied question-answering task because a large language model could confused about whether to directly answer the question or output actions to the target. The supported experiments can be found in the main paper Table 9.

\subsection{Traing Strategy}
We follow the training strategy of NaVid~\cite{zhang2024navid} in a two-stage manner. In the first stage, we firstly pre-train the projector of Uni-NaVid with the Image QA dataset then finetune both the projector and LLM with the Video QA dataset. We collect the data from LLama-Vid~\cite{li2023llama} and Pandm~\cite{chen2024panda}. In the second state, we train both projector and LLM with collected navigation data. The default parameters are borrowed from NaVid~\cite{zhang2024navid}.

\section{Data Preparation Details}

To train \name, we required massive navigation data across different navigation tasks. It is extremely challenging to collect a large number of high-quality annotated data in the real world. Therefore, we collect navigation data in the synthetic environments and we describe the details in the following sections.

\subsection{Data Collection}

\textbf{Navigation samples.} We define the navigation samples including a navigation history video, corresponding instructions, and future actions (we use four actions). Here the navigation history video is accumulated frames to time step $T$, which can be indicated as a set of frames $\{\mathbf{x}_t\}_{1:T}$

\noindent\textbf{Vision-and-language navigation.} We collect VLN navigation samples in the VLN-CE simulator~\cite{Krantz2020BeyondTN, habitat19iccv}. We use the training split of R2R dataset~\cite{Krantz2020BeyondTN} and RxR dataset~\cite{anderson2020rxr}, which include the ground-truth navigation trajectory and corresponding instructions within the HM3D datasets~\cite{ramakrishnan2021hm3d}. Therefore, we deploy the agent to follow the GT trajectory and render RGB images during navigation. In this case, we collect $0.64$ M navigation video-action samples. 

Besides GT navigation samples, we also use DAGGER~\cite{ross2011dagger} to collect more diverse navigation samples, nearly $1.69$ M navigation samples. Specifically, we run \name in the training split and collect the expert actions by using a deterministic path planning method~\cite{sethian1999fast} to the next non-arrived landmark. 

We also collect a sub-split ($70k$) of previously collected VLN data (randomly sample the trajectories which are smaller than $20$ steps) and augment the instruction with low-level actions, \textit{e.g.,} \textit{``move forward 4 steps, then turn right 3 steps."}. We find that \name can easily master the low-level instructions, but the performance drops when the low-level instruction expands significantly.

\noindent\textbf{Object goal navigation.} We collect the object goal navigation data in the HM3D datasets~\cite{ramakrishnan2021hm3d} in Habitat simulator~\cite{habitat19iccv}. Here, the agent is initialized in unseen environments and is required to find one object that could be a category set of {chair, couch, TV, toilet, bed, and plant}. We deploy L3MVN~\cite{yu2023l3mvn} in the HM3D training split and collect the navigation trajectory. We only collect $483$ k navigation trajectory which is successful in finding the target. And we use the instruction template: "\textit{Search for a/an} [Object]." Note that, using the shortest path to the target object as the navigation video samples leads to an extremely low performance ($30.1\%$ SR) because the model can not learn to explore or recover from mistakes.

\noindent\textbf{Embodied question answering.} We collect $240$ k video-action samples from MP3D-EQA dataset~\cite{das2018embodied} and $10$ k video-answering samples. The video-action samples are rendered based on the GT trajectory of MP3D-EQA, and use the question as the instruction. And the video-answering samples are rendered frames of full trajectory, and we use the GT answering as the answer of large language model of MP3D-EQA.

\noindent\textbf{Human following.} We collect $544$ k video-action samples from a self-build human following environment based on the Habitat 3.0 simulator. Specifically, we generate a large amount of human-following data based on the HM3D dataset \cite{ramakrishnan2021hm3d}. For each episode, we first determine the total number of humans (2–6) based on the area of the scene, assign an ID to the target human to be followed, and specify the language instruction. Next, we randomly initialize the starting positions of the humans in the scene and place the robot near the target human, ensuring that the target is visible in the robot’s camera view. Finally, we randomly generate multiple navigation waypoints for each human and use a low-level path planner to guide them to these waypoints sequentially. During this process, the robot continuously receives the target human’s position in each step and uses a local planner to follow the human while avoiding obstacles nearby until the human reaches the final waypoint.

\subsection{Instruction Augmentation}

We collect instructions from various datasets, and the quality of the instructions is limited in diversity. Especially for object goal navigation, the fixed template could cause the instruction to become a task indicator, which could damage the performance. In this case, we use ChatGPT, to augment the instructions. The prompts are listed as follows: 
\begin{tcolorbox}
Given a robot navigation task instruction, rephrase the instruction to make its grammar and structure more diverse while preserving its original meaning. Additionally, ensure all grammatical errors in the instruction are corrected. Use varied sentence structures and descriptions for objects and directions to enhance linguistic diversity. Keep the instructions clear and concise to ensure they remain suitable for a robot navigation system.
\end{tcolorbox}

We find the instruction augmentation could increase the performance of VLN ($+2.31\%$ in SR) and ObjectNav($+3.7\%$ in SR). We believe the cross-dataset instruction augmentation could be a promising topic to further investigate.

\section{Real-world deployment}

\subsection{Robot Setup.}

\begin{figure}
\begin{center}
  \includegraphics[width=0.7 \linewidth]{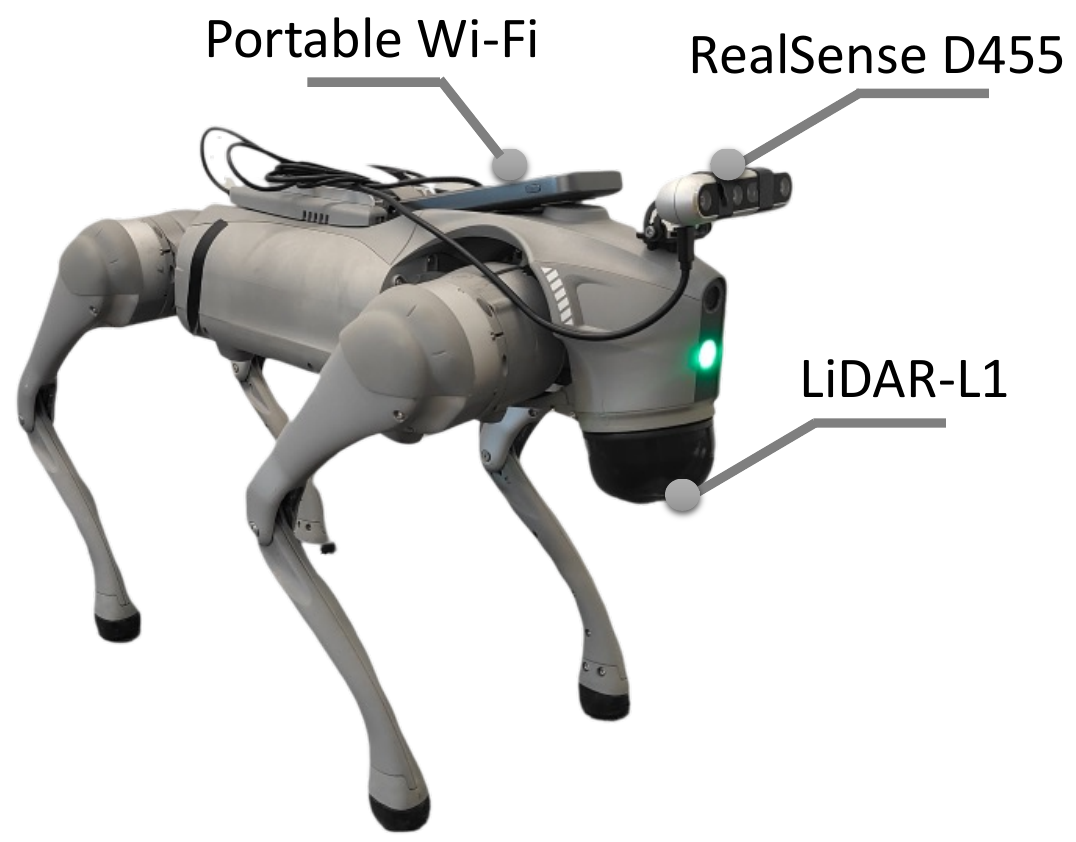}
\end{center}
   \caption{\textbf{Robot setup.} We use Unitree GO2 as our embodiment, and we mount RealSense D455, a portable Wi-Fi and a LiDAR-L1. Note that, our model only takes RGB frames as input. The portable Wi-Fi is used for communication with the remote server and the Lidar is used for the local controller API of Unitree Dog.}
\label{fig:robot_setup}
\end{figure}

We provide a visualization of our robotic dog in Fig.~\ref{fig:robot_setup}. Our robotic dog is Unitree GO2 and we mount a RealSense D455 camera on the head of the robotic dog. Here, we only use the RGB frames with a resolution of $640\times480$ in the setting of  $90^\circ$ HFOV. We also mount a portable Wi-Fi at the back of the robot dog, which is used to communicate with the remote server (send captured images and receive commands). Unitree GO2 is integrated with a LiDAR-L1, which is only used for local motion planning. 

Note that Uni-NaVid does not rely on any odometry algorithms~\cite{zhang2021rosefusion, zhang2022asro} or noiseless depth~\cite{savva2019habitat,zheng2019active, kuang2024openfmnav}, making it easy to deploy in real-world environments.

\subsection{Real-world System Architecture}

\begin{figure}
\begin{center}
  \includegraphics[width=1 \linewidth]{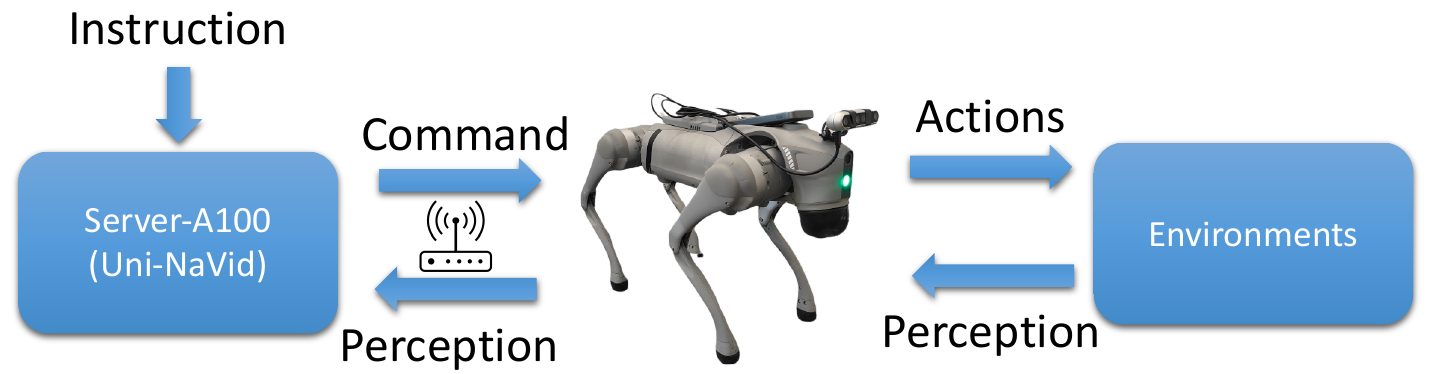}
\end{center}
   \caption{\textbf{Real-world system architecture.} We deploy our model at a remote server, and the robot communicates with the server through the Internet.}
\label{fig:real_world_setup}
\end{figure}

We provide a visualization of the real-world setup in Fig.~\ref{fig:real_world_setup}. Our model is deployed on a remote server equipped with an NVIDIA A100 GPU. During navigation, the server receives navigation instructions and images captured by the robotic dog through the Internet. To ensure efficient communication, the images are compressed before transmission. After processing the newly captured images, the model generates action commands (\texttt{FORWARD}, \texttt{LEFT}, \texttt{RIGHT}, and \texttt{STOP}) and sends them to the robotic dog. Upon receiving these commands, the robotic dog executes the actions using a local motion planning model (specifically, the off-the-shelf model provided by Unitree Dog). Leveraging our online token merging strategy, the model processes newly captured images efficiently, requiring approximately 0.2 seconds for inference, while Internet communication (image transmission and command reception) takes about 0.3 seconds.

\noindent\textbf{Non-blocking navigation.} After receiving a set of four action commands, the robotic dog executes them sequentially. Upon completing each action, the robot captures a new image and transmits it to the server. In cases where multiple action commands are generated for a single step, the robot prioritizes and executes the most recent command, as it reflects the latest planning outcome.

\section{Experiments}

\subsection{Experiment Setup}

\subsubsection{Benchmark.} We conduct extensive experiments on public benchmarks across different tasks. We celebrate them as follows:
\textit{Vision-and-language navigation.} We conduct experiments on the Val-Unseen split of VLN-CE R2R (Main paper Table 2)  and RxR (Main paper Table 3). They contain novel environments and novel instructions. 
\textit{Embodied Question answering.} We conduct experiments on the validation set of the MP3D-EQA~\cite{wijmans2019embodied}, where the questions and scenes are novel. Besides, we also conduct experiments (Table~\ref{tab:comp-eqa}) on OpenEQA~\cite{majumdar2024openeqa}, which is a benchmark to evaluate the performance of understanding the marionettes with egocentric video.
\textit{Human following.} In Table~5 of the main paper, the experiment is conducted on a self-build benchmark based on the HM3D dataset, which also served as the source of the training dataset. Additionally, we also build two benchmarks based on the HSSD dataset and MP3D dataset, respectively, and evaluate the human following capability of \name~across different datasets and scenes (Table~\ref{tab:human-following-cross}).
    

\subsubsection{Metrics.} 

We use the default metrics of each benchmark. For vision-and-language navigation and object goal navigation,~\cite{Krantz2020BeyondTN,savva2019habitat,wu2024embodied}, we use SR, OSR, and SPL as metrics. Specifically, SR (Success Rate) measures the proportion of tasks in which the agent successfully reaches the target location (in VLN the distance threshold is $3$ m, and in ObjectNav is $1$ m) within the allowed time steps (Up to 500 steps). OSR (Oracle Success Rate) extends SR by accounting for the agent’s proximity to the goal, considering the task successful if the agent is close enough, even without directly stopping at the target. SPL (Success weighted by Path Length) evaluates the agent’s efficiency by combining success with the optimality of its trajectory, penalizing longer-than-necessary paths while rewarding those closer to the shortest possible route.  

For human following navigation, we use SR, FR, and CR metrics. SR (Success Rate) measures the proportion of events in which the agent successfully follows the target human to the endpoint. FR (Following Rate) refers to the proportion of steps in which the agent successfully follows the target human for each step. CR (Collision Rate) refers to the proportion of episodes in which the agent collides with the human during the movement.

For embodied question answering, we use ACC metric, which directly measures the percentage of correct answers. For ScnaQA~\cite{azuma2022scanqa}, we use EM, BLUE, ROUGE, METEOR, CIDEr as metrics. Specifically, EM (Exact Match) evaluates the percentage of predictions that exactly match the reference answers. BLEU (Bilingual Evaluation Understudy) measures the n-gram overlap between the generated text and the reference, rewarding precise matches in shorter sequences. ROUGE (Recall-Oriented Understudy for Gisting Evaluation) assesses the overlap of word sequences, focusing on recall and capturing the informativeness of the generated text. METEOR (Metric for Evaluation of Translation with Explicit ORdering) combines precision and recall, using synonym matching and stemming to account for variations in phrasing. Lastly, CIDEr (Consensus-based Image Description Evaluation) measures the similarity of generated responses to multiple references by capturing human consensus, emphasizing relevance and diversity in the output.

For video-question answering benchmark MSVD-QA~\cite{xu2017video}, MSRVTT-QA~\cite{xu2017video}, and ActivityNet-QA~\cite{yu2019activitynet}, we use ACC and Score metrics. Specifically, ACC (Accuracy) measures the proportion of correctly answered questions, providing a straightforward assessment of the model’s overall correctness. Score, on the other hand, evaluates the quality of the generated answers using GPT (GPT 3.5 in implementation) as the zero-shot evaluation assistor to assign relative scores on a scale of 1 to 5 for generated answers.


\subsection{Real-world Experiments}

We conduct real-world experiments to study the generalizability of our method in the real world. Specifically, we leverage the VLN task, which includes both landmarks and motions, to evaluate our method with the previous VLN method NaVid~\cite{zhang2024navid}. Following previous work~\cite{zhang2024navid}, we designed two types of instructions for different difficulties (25 simple instructions and 25 complex instructions). The simple instructions, which require the agent to navigate to a single robot landmark and stop; (2) complex instructions, which require the agent to follow a series of simple instructions.

We list some examples of used instructions in experiments. Simple instruction examples:

\begin{itemize}
    \item Move forward 1 step, then stop.
    \item Turn left 5 steps, then stop.
    \item Move to the right chair, then turn left.
\end{itemize}

Complex instructions:

\begin{itemize}
    \item Move forward 5 steps, then turn left 4 steps, and finally turn right 5 steps.
    \item Go straight to the chair, then turn right and move to the door, stop by the door.
    \item  Turn a large right, and go forward to a plant, then right at the plant and move to the TV, stop close to TV.
\end{itemize}

\begin{table}[!t]
\centering

\scalebox{0.9}{
\setlength{\tabcolsep}{0.9mm}{
\begin{tabular}{lll|cc}
\hline
\multicolumn{3}{l|}{Method}    & Simple Ins. & Complex Ins. \\ \hline
\multicolumn{3}{l|}{NaVid~\cite{zhang2024navid}}     & 80$\%$         & 20$\%$           \\
\multicolumn{3}{l|}{Uni-NaVid} & 92$\%$          & 84$\%$           \\ \hline
\end{tabular}
}
}
\caption{\textbf{Real-world VLN experiments.} We compare our method with NaVid on two types of instructions: simple instructions (25) and complex instructions (25).}
\label{tab:comp-real-world}
\vspace{-3mm}
\end{table}

Here we present our results in Table~\ref{tab:comp-real-world}. We find that both Navid and our method can achieve high SR in simple instructions, However, for complex instructions, our method shows significant improvements. This demonstrates the superiority of our method over existing methods. We add more visual results to the attached video.

\subsection{More Benchmark Experiments}

\begin{table}[!t]
\centering

\resizebox{\linewidth}{!}{
\scalebox{1}{
\setlength{\tabcolsep}{0.8mm}{
\begin{tabular}{l|ccc|ccccc}
\hline
\multirow{2}{*}{Method}         & \multicolumn{3}{c|}{Observation}     & \multicolumn{5}{c}{VLN-CE RxR Val-Unseen}                             \\ 
 & Odom. & Depth & S.RGB & TL & \textbf{NE}$\downarrow$ & \textbf{OS}$\uparrow$ & \textbf{SR}$\uparrow$ & \textbf{SPL}$\uparrow$ \\ \hline
LAW~\cite{raychaudhuri2021law}  & \checkmark & \checkmark & \checkmark & 4.01  & 10.87         & 21.0          & 8.0           & 8.0           \\
CM2~\cite{georgakis2022cross}   & \checkmark & \checkmark & \checkmark & 12.29 & 8.98          & 25.3          & 14.4          & 9.2           \\
WS-MGMap~\cite{chen2022weakly}  & \checkmark & \checkmark & \checkmark & 10.80 & 9.83          & 29.8          & 15.0          & 12.1          \\
ETPNav.FF~\cite{wang2024sim}                       & \checkmark & \checkmark & \checkmark & -     & 8.79          & 36.7          & 25.5          & 18.1          \\
Seq2Seq~\cite{Krantz2020BeyondTN} &            & \checkmark & \checkmark & 1.16  & 11.8          & 5.02          & 3.51          & 3.43          \\
CMA~\cite{Krantz2020BeyondTN}     &            & \checkmark & \checkmark & 5.09  & 11.7          & 10.7          & 4.41          & 2.47          \\
$A^2$Nav$^\dagger$~\cite{chen20232}       &            &            & \checkmark & --    & --            & --            & 16.8          & 6.3           \\
NaVid~\cite{zhang2024navid}                          &            &            & \checkmark & 10.59 & 8.41          & 34.5          & 23.8          & 21.2          \\
\textbf{Uni-NaVid}                       &            &            & \checkmark & 8.3  & \textbf{8.08} & \textbf{40.9} & \textbf{29.5} & \textbf{28.1} \\ \hline
\end{tabular}
}
}
}
\caption{\textbf{Vision-and-language navigation (RxR).} Comparison on VLN-CE RxR~\cite{ku2020room} Val-Unseen. $^\dagger$: indicates zero-shot methods.}

\label{tab:comp-vlnce-rxr-cross-dataset}
\end{table}


\noindent\textbf{Cross-dataset Vision-and-Language Navigation Evaluation.} We evaluate the cross-dataset performance of our method on VLN-CE RxR by excluding the VLN-CE RxR data at training time. The results are presented in Table~\ref{tab:comp-vlnce-rxr-cross-dataset}. Notably, even without training on RxR, our method still outperforms existing approaches. However, a significant decrease in all metrics is observed compared to when RxR training data is included. We hypothesize that this discrepancy arises from differences in trajectory characteristics: trajectories in the R2R dataset have relatively uniform lengths (approximately 10 meters), whereas those in RxR exhibit greater diversity, ranging from approximately 2 meters to 20 meters. This disparity constrains our method’s performance on the RxR dataset and underscores the importance of training on diverse trajectories.



\begin{table}[!t]

\centering

\scalebox{0.78}{
\setlength{\tabcolsep}{0.8mm}{
\begin{tabular}{lccc}
\hline
\multirow{3}{*}{\large Method} & \multicolumn{3}{c}{EM-EQA}                                                                                                 \\
                        & \makecell{ScanNet \\ Eq. (1)} & \makecell{HM3D \\ Eq. (1)}  & \makecell{ALL \\ Eq. (1)} \\ \hline
\multicolumn{4}{l}{\colorbox{green!20}{\large Blind LLMs}} \\
GPT-4 & 32.5 & 35.5 & 33.5 \\
LLaMA-2 & 27.9 & 29.0 & 28.3 \\ \hline
\multicolumn{4}{l}{\colorbox{green!20}{Socratic LLMs w/ Frame Captions}} \\
GPT-4 w/ LLaVA-1.5 & 45.4 & 40.0 & 43.6 \\
LLaMA-2 w/ LLaVA-1.5 & 39.6 & 31.1 & 36.8 \\ \hline
\multicolumn{4}{l}{\colorbox{green!20}{Socratic LLMs w/ Scene-Graph Captions}} \\
GPT-4 w/ CG & 37.8 & 34.0 & 36.5 \\ 
LLaMA-2 w/ CG & 31.0 & 24.2 & 28.7 \\
GPT-4 w/ SVM & 40.9 & 35.0 & 38.9 \\
LLaMA-2 w/ SVM & 36.0 & 30.9 & 34.3 \\ \hline
\multicolumn{4}{l}{\colorbox{green!20}{Multi-Frame VLMs}} \\ 
GPT-4V & 51.3 & 46.6 & 49.6 \\
Human Agent & 87.7 & 85.1 & 86.8 \\ \hline
\name & 41.4 & 38.1 & 40.3 \\ \hline
\end{tabular}
}
}
\caption{\textbf{Embodied video question answering.} Comparison on OpenEQA benchmark.  EM-EQA results are broken down by data source (ScanNet, HM3D,
and ALL). GPT-4V scores are calculated on a subset of 500 OpenEQA questions due to API limitations.}
\label{tab:comp-eqa}
\end{table}

\noindent\textbf{OpenEQA benchmark~\cite{majumdar2024openeqa}.} To further evaluate our method on embodied question answering, we conduct experiments in OpenEQA, which require the methods to answer questions by analyzing an egocentric video. The results are shown in Tab.~\ref{tab:comp-eqa}. We find our method achieves competitive performance to the strongest commercial used Vision-Language model such as GPT-4V. Besides, by directly observing and understanding the video sequences, our method does not need additional frame caption or Socratic-style promoted~\cite{majumdar2024openeqa} used in other LLMs.

\begin{table}[!t]
\centering

\scalebox{0.8}{
\setlength{\tabcolsep}{1mm}{
\begin{tabular}{l|ccc|ccc}
\hline
\multirow{2}{*}{Method} & \multicolumn{3}{c|}{HF-HSSD} & \multicolumn{3}{c}{HF-MP3D}\\ &
\hspace{0.3cm}\textbf{SR}$\uparrow$\hspace{0.2cm} & \hspace{0.3cm}\textbf{FR}$\uparrow$\hspace{0.3cm} & \hspace{0.1cm}\textbf{CR}$\downarrow$ & \hspace{0.3cm}\textbf{SR}$\uparrow$\hspace{0.2cm} & \hspace{0.3cm}\textbf{FR}$\uparrow$\hspace{0.3cm} & \hspace{0.1cm}\textbf{CR}$\downarrow$\\
\hline

PoliFormer~\cite{zeng2024poliformer} & 2.67 & 20.81 & 0.97 & 2.62 & 16.59 & 1.43 \\
PoliFormer$^*$~\cite{zeng2024poliformer} & 26.97 & 54.20 & 10.01 & 25.42 & 47.80 & 8.72 \\
PoliFormer$\dag $~\cite{zeng2024poliformer}      & 27.10 & 53.96 & 9.34 & 21.90 & 41.42 & 7.15 \\
IBVS$^*$~\cite{gupta2016novel} & 66.32 & 80.26 & 0.19 
& 56.86 & 68.91 & 1.33\\
IBVS$\dag $~\cite{gupta2016novel} & 65.36 & 80.33 & \textbf{0.15} & 58.15 & 65.83 & \textbf{0.77} \\
\textbf{\name}    & \textbf{81.65} & \textbf{89.34} & 1.33 & \textbf{69.80} & \textbf{78.96} & 2.99 \\ \hline
\end{tabular}
}
}
\caption{\textbf{Human following.} Comparison on Human Following HSSD Benchmark (HF-HSSD) and Human Following MP3D Benchmark (HF-MP3D). $^*$: Methods use GroundingDINO~\cite{liu2023grounding} as the open-vocabulary human detector. $\dag$: Methods use the ground-truth bounding box provided by the simulator.}
\label{tab:human-following-cross}
\end{table}

\noindent\textbf{Cross-Environments Following.} We evaluate our method in novel environments, including HSSD~\cite{khanna2024habitat} (synthetic environments) and MP3D~\cite{chang2017matterport3d} (reconstructed environments). The results, presented in Table~\ref{tab:human-following-cross}, demonstrate that our approach consistently outperforms baseline methods in both SR and FR metrics across HSSD and MP3D. Notably, our method achieves significant improvements in HSSD, likely due to the absence of reconstruction artifacts in synthetic environments, which reduces the likelihood of the robot being stuck. For the CR (collision rate) metric, IBVS adopts a highly conservative strategy that maintains a considerable distance from the target (i.e., a large bounding box). While this results in a lower CR, it adversely affects FR and SR performance.

\subsection{More Ablation study}

\begin{table}[!t]
\centering

\scalebox{0.7}{
\setlength{\tabcolsep}{0.72mm}{
\begin{tabular}{lll|cccc}
\hline
\multicolumn{3}{l|}{Design} & VLN (SR$\uparrow$) & ObjNav (SR$\uparrow$) & EQA (ACC$\uparrow$) & Follow (SR$\uparrow$) \\ \hline
\multicolumn{3}{l|}{4 tokens for Curr. Obs.} & 43.4 & 50.2 & 40.5 & -    \\
\multicolumn{3}{l|}{1 token for Curr. Obs.}  & 35.1 & 45.3 & 31.2 & -    \\
\multicolumn{3}{l|}{1 token for Short. Obs.} & 46.2 & 69.2 & 44.9 & 60.3 \\ \hline
\multicolumn{3}{l|}{$\tau$=0.9}              & 40.2 & 70.0 & 43.2 & 57.8 \\
\multicolumn{3}{l|}{$\tau$=0.95}               & 48.7 & 73.7 & 47.3 & 61.2 \\
\multicolumn{3}{l|}{$\tau$=0.99}               & 49.6 & 70.2 & 48.1 & 57.2 \\ \hline
\end{tabular}
}
}
\caption{\textbf{Ablation study} of token number and $\tau$}
\label{tab:ablation-study-suppl}
\end{table}



We conduct additional ablation studies to validate the effectiveness of our key design components. The results are presented in Table~\ref{tab:ablation-study-suppl}. Notably, we observed a significant performance drop when the number of current observation tokens was reduced. In particular, the following task failed across all sequences, potentially due to the inability of a smaller number of tokens to provide sufficient information for tracking human motion. Furthermore, we found that increasing the value of $\tau$ led to improved performance. This outcome is intuitive, as a higher $\tau$ retains more long-term observation tokens, which are crucial for understanding navigation history.


\subsection{Time Analyze}

\begin{figure}
\begin{center}
  \includegraphics[width=1 \linewidth]{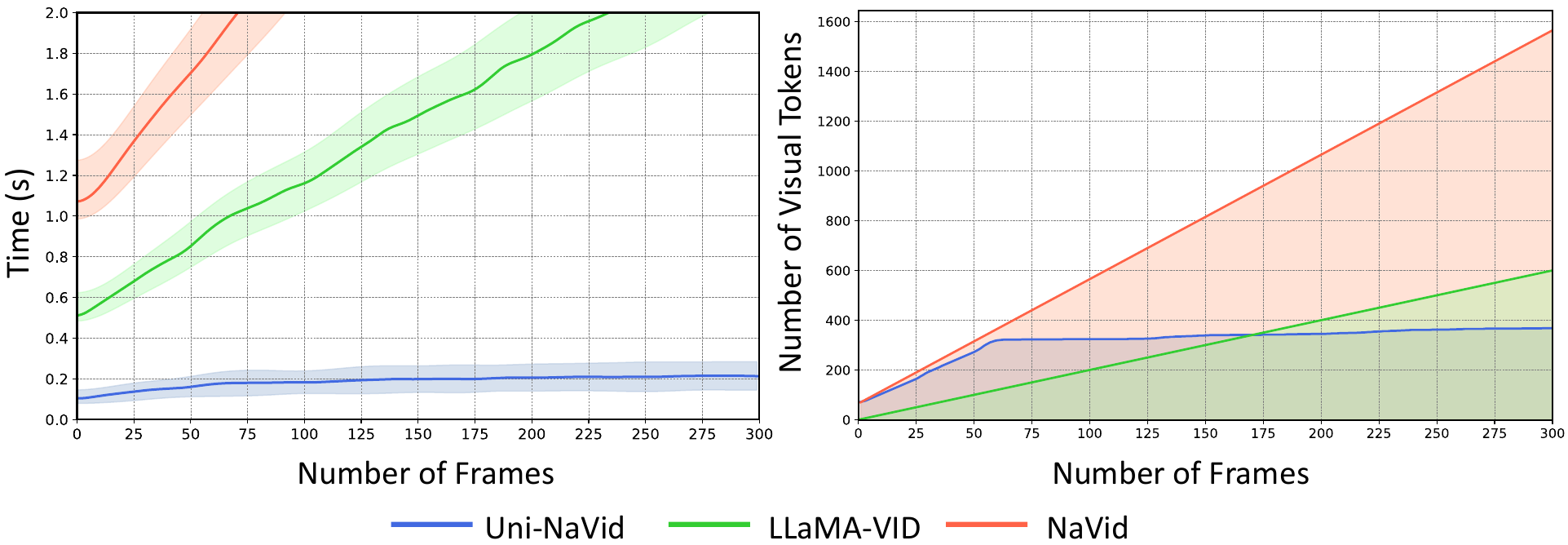}
\end{center}
   \caption{\textbf{Time efficiency visualization.} We provide the average running time and the number of tokens across different time steps.}
\label{fig:time_analyze}
\end{figure}

To evaluate the efficiency of our method, we present the average running time and token count across different time steps in Fig.~\ref{fig:time_analyze}. Our approach is compared against existing models that also employ token merging strategies. The results demonstrate that our method is more efficient which achieves an inference time of approximately $0.2$ seconds. Moreover, our method maintains consistent running times, whereas the running times of existing methods increase significantly. These advantages stem from two key factors: \textit{First}, our model architecture is highly efficient, whereas other approaches, such as LLaMA-VID~\cite{li2023llama} and NaVid~\cite{zhang2024navid}, rely on time-consuming operations like QFormer. \textit{Second}, our online token merging strategy results in a gradual increase in token count, ensuring more stable inference times.

\subsection{Qualitative Experiments}

\begin{figure*}
\begin{center}
  \includegraphics[width=1 \linewidth]{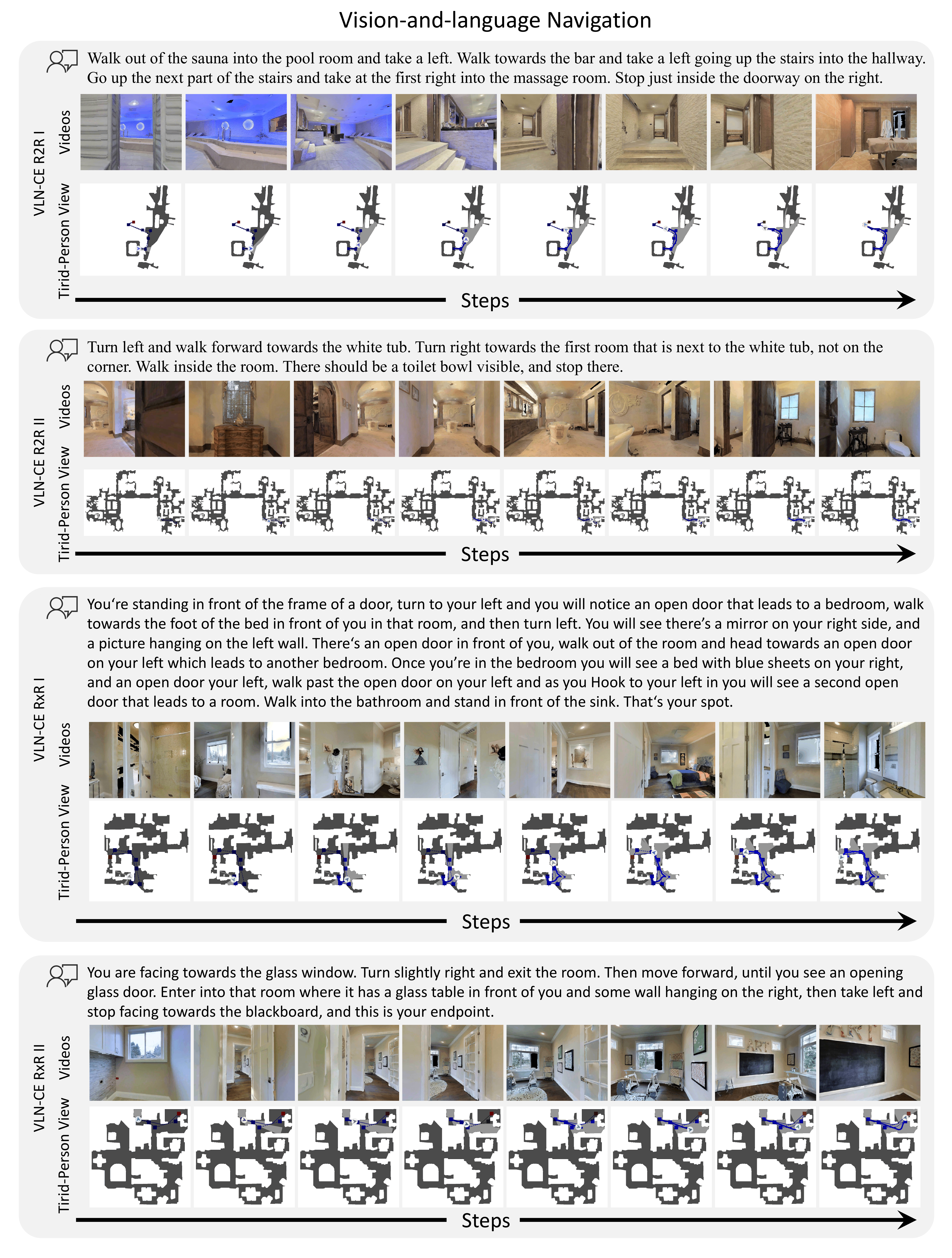}
\end{center}
   \caption{Visual results of VLN on VLN-CE R2R and RxR.}
\label{fig:gallery_vln}
\end{figure*}

\begin{figure*}
\begin{center}
  \includegraphics[width=1 \linewidth]{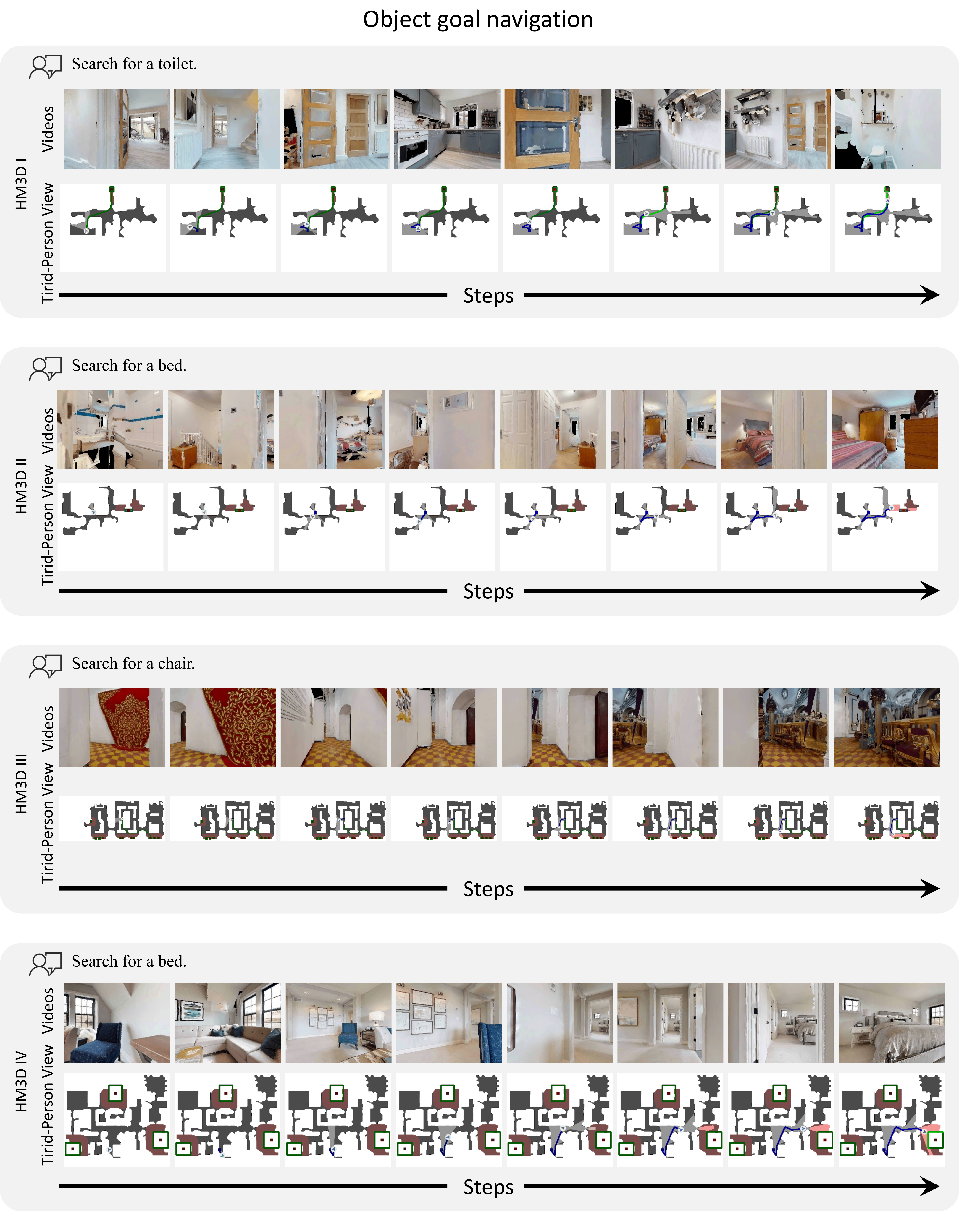}
\end{center}
   \caption{Visual results of HM3D ObjectNav.}
\label{fig:gallery_objnav}
\end{figure*}

\begin{figure*}
\begin{center}
  \includegraphics[width=1 \linewidth]{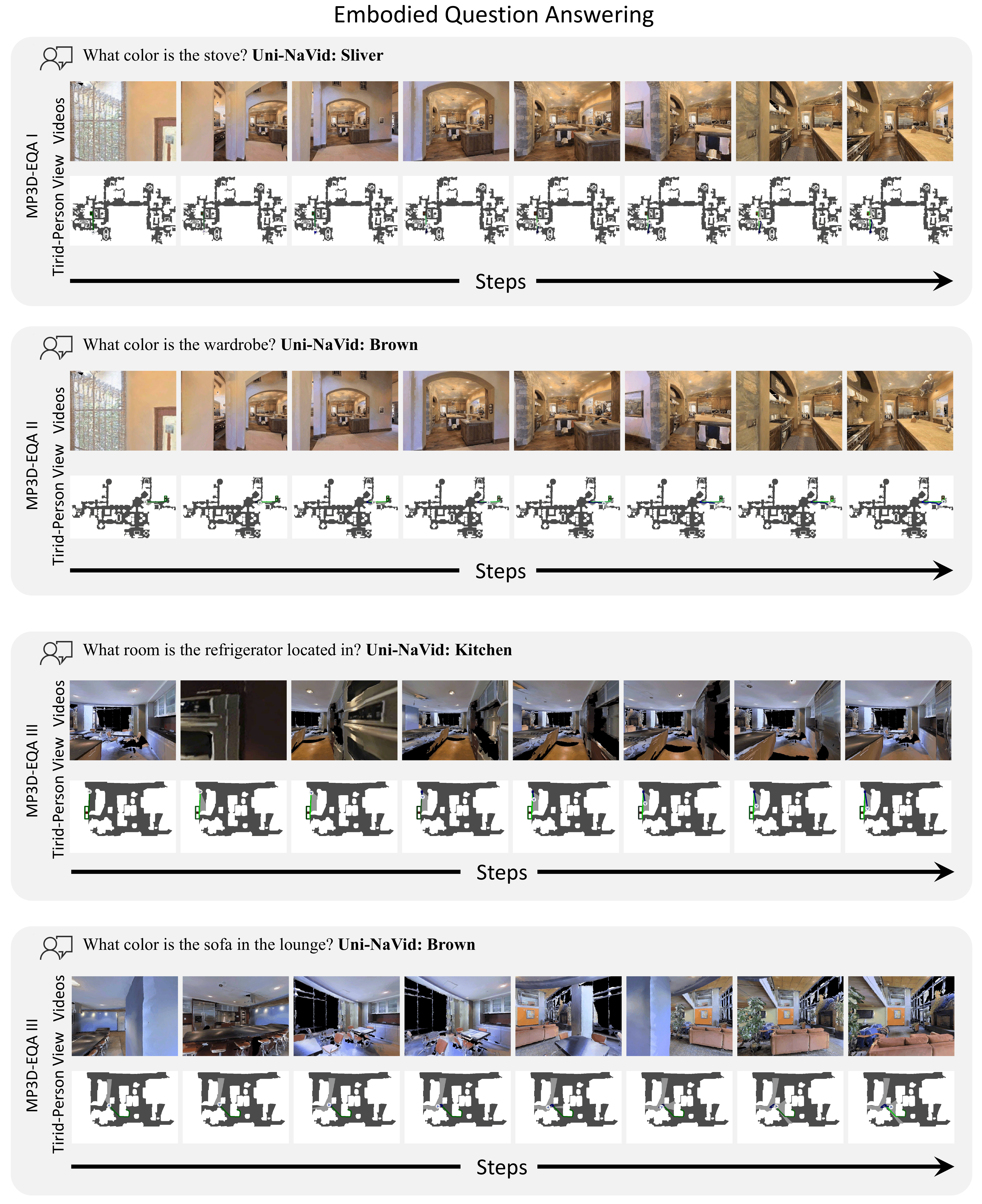}
\end{center}
   \caption{Visual results of MP3D-EQA.}
\label{fig:gallery_eqa}
\end{figure*}

\begin{figure*}
\begin{center}
  \includegraphics[width=1 \linewidth]{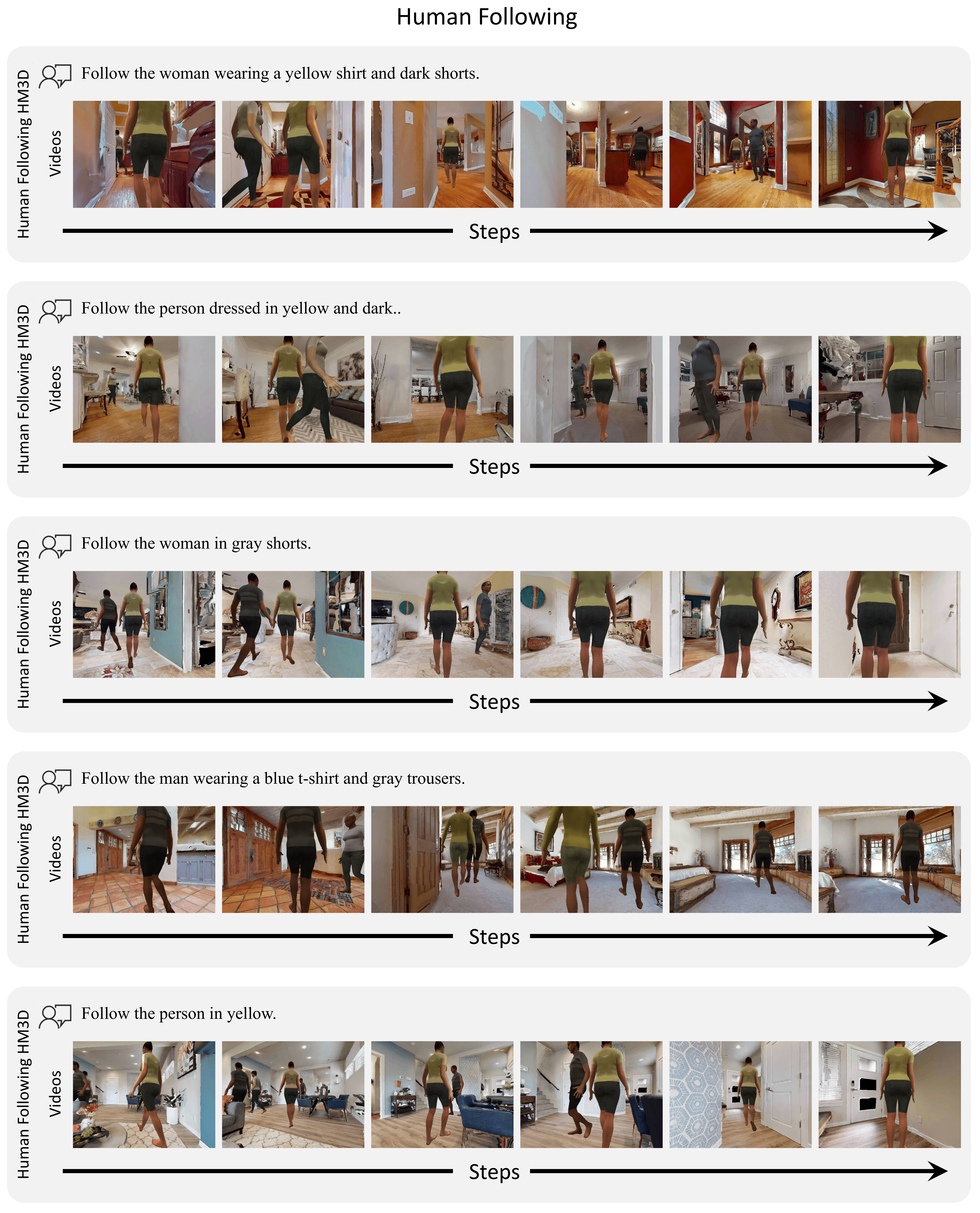}
\end{center}
   \caption{Visual results of HM3D Human following.}
\label{fig:gallery_hf}
\end{figure*}

We provide visual results of our method on various benchmarks: Fig.\ref{fig:gallery_vln} illustrates results for VLN-CE R2R and RxR, Fig.\ref{fig:gallery_objnav} for HM3D ObjectNav, and Fig.~\ref{fig:gallery_eqa} for MP3D-EQA, Fig.~\ref{fig:gallery_hf} for HM3D human following. For additional visual results, please refer to the attached video.

\end{document}